\documentclass{article} 
\usepackage[final]{colm2026_conference}

\usepackage{microtype}
\usepackage{hyperref}
\usepackage{url}
\usepackage{booktabs}
\usepackage{enumitem}
\usepackage{amsmath}
\usepackage{amssymb}
\usepackage{mathtools}
\usepackage{amsthm}
\usepackage{etoc}

\usepackage{placeins}
\usepackage{tcolorbox}
\tcbuselibrary{breakable}
\usepackage{booktabs}
\usepackage{bbding}
\usepackage{cleveref}
\usepackage{graphicx}
\definecolor{darkmagenta}{rgb}{0.56, 0.0, 1.0} 
\usepackage{multirow}
\usepackage{inconsolata}
\usepackage{changepage}
\usepackage{newtxmath}
\usepackage{cleveref}
\usepackage{xcolor}
\usepackage{mathtools}
\usepackage{bbm}
\usepackage{xspace}
\usepackage{wrapfig}
\usepackage{wrapfig}
\usepackage{caption}

\usepackage{lineno}

\definecolor{darkblue}{rgb}{0, 0, 0.5}
\hypersetup{colorlinks=true, citecolor=darkblue, linkcolor=darkblue, urlcolor=darkblue}

\hypersetup{colorlinks,linkcolor={orange},citecolor={darkmagenta},urlcolor={orange}} 
\hypersetup{colorlinks,linkcolor={blue},citecolor={darkmagenta},urlcolor={blue}} 

\title{A Comedy of Estimators: \\On KL Regularization in RL Training of LLMs
}

\author{\textbf{Vedant Shah$^{*\;1,2}$, Johan Obando-Ceron$^{*\;1,2}$, Vineet Jain$^{*\;1,3}$} \\
\textbf{Brian R. Bartoldson$^{5}$, Bhavya Kailkhura$^{5}$, Sarthak Mittal$^{1,2}$, Glen Berseth$^{1,2,7}$} \\
\textbf{Pablo Samuel Castro$^{1,2}$, Yoshua Bengio$^{1,2,4,7}$, Esmeralda S. Whitammer$^{6,8}$} \\
\textbf{Moksh Jain$^{\dagger\;1,2}$, Siddarth Venkatraman$^{\dagger\;1,2}$, Aaron Courville$^{\dagger\;1,2,7}$} \\
$^{*}$Equal contribution \quad
{$\dagger$} Equal senior contribution\\
\textsuperscript{1}Mila -- Qu\'ebec AI Institute
\quad\textsuperscript{2}Universit\'e de Montr\'eal
\quad\textsuperscript{3}McGill University \\
\textsuperscript{4}LawZero
\quad\textsuperscript{5}LLNL
\quad\textsuperscript{6}University of Edinburgh
\quad\textsuperscript{7}CIFAR AI Chair
\quad\textsuperscript{8}CIFAR Fellow\\
$\left\{
    \begin{array}{@{}l@{}}
    \text{vedant.shah, johan.ceron, jain.vineet, mittalsa, glen.berseth,}\\
    \text{pablo-samuel.castro, yoshua.bengio, moksh.jain,}\\
    \text{siddarth.venkatraman, courvila}
  \end{array}
\right\}$@mila.quebec\\ \\
\{bartoldson1, kailkhura1\}@llnl.gov, esmeralda.whitammer@ed.ac.uk%
}

\usepackage{amsmath,amsfonts,bm}

\def\eqref#1{equation~\ref{#1}}

\def\1{\bm{1}}

\DeclareMathAlphabet{\mathsfit}{\encodingdefault}{\sfdefault}{m}{sl}
\SetMathAlphabet{\mathsfit}{bold}{\encodingdefault}{\sfdefault}{bx}{n}

\def\gD{{\mathcal{D}}}

\newcommand{\E}{\mathbb{E}}

\newcommand{\KL}{D_{\mathrm{KL}}}

\usepackage{xcolor}
\definecolor{myblue}{RGB}{0,80,200}

\usepackage{soul}
\definecolor{pastelMedium}{RGB}{244, 196, 200}

\usepackage{soul}
\definecolor{dullGreen}{RGB}{153, 187, 153}

\definecolor{pastelSmall}{RGB}{190, 215, 255} 

\usepackage{soul}
\definecolor{darkGreen}{RGB}{0, 100, 0}

\def\pith{{\pi_\theta}}
\def\piref{{\pi_\text{ref}}}
\def\kl{{\mathbb{KL}}}
\def\yseq{{y_{1:T}}}
\def\yt{{y_t}}
\def\hist{{x, y_{<t}}}
\def\d{{\nabla_{\theta}}}
\def\klest{{\widehat{\mathrm{KL}}}}
\def\expec{{\mathbb{E}_{\yseq \sim \pith(\cdot \mid x)}}}
\newcommand{\ie}{\textit{i.e.}}
\newcommand{\eg}{\textit{e.g.}}

\definecolor{varC}{RGB}{160, 120, 0}
\definecolor{fRed}{RGB}{158, 12, 9}
\definecolor{fBlue}{RGB}{25, 117, 197}

\newcommand{\naivet}{%
  \hyperref[eq:naive]{\textcolor{fRed}{\texttt{K1}_t}}%
}
\newcommand{\naive}{%
  \hyperref[eq:naive]{\textcolor{fRed}{\texttt{K1}}}%
}

\newcommand{\lowvart}{%
  \hyperref[eq:lowvar]{\textcolor{fBlue}{\texttt{K3}_t}}%
  \xspace
}
\newcommand{\lowvar}{%
  \hyperref[eq:lowvar]{\textcolor{fBlue}{\texttt{K3}}}%
  \xspace
}

\renewcommand{\sectionautorefname}{Sec.}
\renewcommand{\subsectionautorefname}{Sec.}
\renewcommand{\subsubsectionautorefname}{Sec.}

\makeatletter
\renewcommand*{\sectionautorefname}{\S\@gobble}
\renewcommand*{\subsectionautorefname}{\S\@gobble}
\renewcommand*{\subsubsectionautorefname}{\S\@gobble}

\Crefname{equation}{Eq.}{Eqs.}
\crefname{equation}{Eq.}{Eqs.}

\Crefname{figure}{Fig.}{Figs.}
\crefname{figure}{Fig.}{Figs.}

\Crefname{algorithm}{Alg.}{Algs.}
\crefname{algorithm}{Alg.}{Algs.}

\Crefname{appendix}{App.}{Apps.}
\crefname{appendix}{App.}{Apps.}

\Crefname{section}{\S\@gobble}{\S\@gobble}
\crefname{section}{\S\@gobble}{\S\@gobble}
\Crefname{subsection}{\S\@gobble}{\S\@gobble}
\crefname{subsection}{\S\@gobble}{\S\@gobble}
\Crefname{subsubsection}{\S\@gobble}{\S\@gobble}
\crefname{subsubsection}{\S\@gobble}{\S\@gobble}

\makeatletter
\def\addcontentsline#1#2#3{%
\addtocontents{#1}{\protect\contentsline{#2}{#3}{\thepage}{\@currentHref}{}}}
\makeatother

\begin{document}
\etocdepthtag.toc{mtoc}
\etocsettagdepth{mtoc}{none}
\etocsettagdepth{appendix}{none}

\ifcolmsubmission
\linenumbers
\fi

\maketitle

\begin{abstract}
The reverse Kullback-Leibler (KL) divergence between the trained policy and a reference policy is commonly used as a regularizer when training large language models (LLMs) with reinforcement learning (RL). Since computing the exact sequence-level KL divergence is intractable, practical algorithms use sample-based estimators computed from on-policy rollouts and incorporate them in the reward or the loss. 
Despite its ubiquity, the interaction between estimator choice, position of the regularization, and downstream performance is poorly understood. Recent work also identifies that certain implementation choices can result in biased gradients. We analyze these practices and examine the gradients of several estimator configurations, showing how the choices shape gradient bias. We substantiate these observations with empirical evidence by RL fine-tuning LLMs with different KL configurations, and evaluating their performance on both in- and out-of-distribution tasks.
In on-policy settings, we find that biased-gradient estimator configurations can show training instabilities, while unbiased-gradient configurations lead to better performance on in-domain as well as out-of-domain tasks. We also observe that KL regularization can help stabilize asynchronous training. Overall, our findings provide useful takeaways for KL-regularized objectives during RL post-training of LLMs.
\end{abstract}

\section{Introduction}

Reinforcement learning (RL) is a key component of post-training pipelines for large language models (LLMs). 
\citet{jaech2024openai,guo2025deepseek} have shown that training LLMs with RL on verifiable reasoning tasks such as math and coding improves reasoning abilities. This has precipitated rapid development in reasoning-oriented RL approaches for LLMs~\citep{rlhf2024}. At the same time, this has resulted in inconsistent design choices and implementations in open-source RL fine-tuning pipelines~\citep{tang2025few,zhang2025design}.

One such design choice is the use of Kullback-Leibler (KL) divergence between the trained policy and a reference policy as a regularization term in the objective~\citep{ziegler2019fine,shao2024deepseekmath}. 
It is designed to constrain the policy within the support of the reference model, mitigating problems such as language drift, reward over-optimization~\citep{gao2023scaling}, and catastrophic forgetting~\citep{qi2024online}. Traditionally, the \emph{reverse} KL divergence is used for this regularization, with 
a regularization coefficient controlling the trade-off between reward maximization and proximity to the reference model. However, exact computation of the reverse KL divergence is intractable due to the combinatorially large space of possible sequences. As a result, different sample-based estimators of the reverse KL divergence are used in practice~\citep{schulman2020kl,amini2025better}.

In addition to differences in their approximations, these estimators may be incorporated into the objective in different ways: RLHF with PPO~\citep{ouyang2022training} adds the KL penalty to the task reward (\ie, no direct gradients); methods such as GRPO~\citep{shao2024deepseekmath,guo2025deepseek} popularized adding it directly to the loss. The choice of the estimator, regularization coefficient, and whether it is added to the reward or directly to the loss can have a significant effect on the training stability, convergence rate, and out-of-distribution generalization of the trained models. Moreover, previous work \citep{tang2025few,zhang2025design} has identified that some of these choices lead to biased estimates of the true gradient. For example, using the KL estimator in the loss function, as popularized by GRPO, results in biased gradients and therefore does not optimize the intended reverse KL-regularized objective. 



While KL regularization is ubiquitous in RL training of LLMs and theoretically well-understood, there is a lack of empirical investigation of the effects of these choices on the downstream performance. \textbf{In this work, we fill this gap by providing a systematic analysis and empirical investigation of the space of these design choices, with respect to \textit{sequence-level reverse} KL regularization as the intended objective}. We study this in the context of reinforcement learning with verifiable rewards~\citep[RLVR;][]{trung2024reft,lambert2025tulu}, which has become the dominant paradigm for improving the reasoning abilities of LLMs. 
Specifically, we investigate two commonly used \emph{unbiased} estimators of reverse KL divergence in on-policy settings -- the naïve or $\naive$ estimator, and the Schulman or \lowvar estimator~\citep{schulman2020kl}. These are the most common estimators used in practice, hence we limit our scope to these estimators and do not include others, such as the squared or \texttt{K2} estimator, in our analysis.

First, we give an overview of the bias of the gradients induced by different estimator configurations with respect to the true gradient of \textit{sequence-level} reverse KL divergence. While both estimators are unbiased, they do not necessarily induce an unbiased gradient update
(\Cref{sec:estimators}, \cref{tab:overview}). We demonstrate this empirically in a minimal setting where computing the true gradient of sequence-level reverse KL divergence is tractable (\cref{sec:toy}). Next, we empirically study the implications of these choices in RL fine-tuning of \texttt{Qwen2.5-7B}~\citep{Yang2024Qwen25TR} and \texttt{Llama-3.1-8B-Instruct}~\citep{touvron2023llama} on a mathematical reasoning task in an on-policy setting. We also study both in- and out-of-domain performance of the resulting models (\cref{sec:results}). Finally, we conduct a similar study in a highly asynchronous RL training setup~\citep{noukhovitch2024asynchronous, bartoldson2025trajectory} by training \texttt{Qwen2.5-7B} on a mathematical reasoning task and \texttt{Qwen3-4B-Instruct-2507}~\citep{yang2025qwen3} on a general reasoning task. 

\begin{tcolorbox}[colback=orange!10,
leftrule=0.5mm,top=1mm,bottom=1mm,boxrule=0.6pt,breakable]
\textbf{Key observations}:
\begin{itemize}[left=0cm,nosep]
    \item Unbiased estimates of the sequence-level reverse KL divergence can result in \emph{biased} gradients depending on their usage.
    \item Configurations inducing biased gradients can lead to unstable training and even precipitate collapse.
    \item Configurations that lead to unbiased gradient estimates result in better-performing models, on both in- and out-of-domain evaluation tasks.
    \item KL regularization can stabilize learning from highly off-policy data in asynchronous setups.
\end{itemize}
\end{tcolorbox}

\section{Background}
We study the problem of fine-tuning a language model $\pi_{\theta}$ with reinforcement learning. Given a reward function $R(\cdot)$ and a set of observations $\gD$ comprising question-answer pairs $(x, y)$, RL fine-tuning of LLMs optimizes the following objective
\begin{equation}
\label{eq:rl-training}
\begin{aligned}
\max_\theta\;\mathbb{E}_{(x,y)\sim\mathcal{D}}\Big[
\mathbb{E}_{\yseq\sim \pi_\theta(\cdot\mid x)}\big[R(y_{1:T},y)\big]
-\beta\,\mathrm{KL}\!\left(\pi_\theta(\cdot\mid x)\,\|\,\pi_{\mathrm{ref}}(\cdot\mid x)\right)
\Big],
\end{aligned}
\end{equation}
where $\beta$ is a hyperparameter that controls the weight of the KL divergence penalty, $\pi_{\mathrm{ref}}$ is the reference policy, and $\yseq$ denotes solutions generated by the model conditioned on the question. Since both the sampling of $\yseq$ and $R$ are non-differentiable, the objective is optimized using policy gradient methods such as PPO~\citep{ouyang2022training} or GRPO~\citep{shao2024deepseekmath}.

We study the reinforcement learning with verifiable rewards (RLVR) setting where $R$ is an oracle verifier. An example is a reward function that checks the generated answer against the ground truth answer: $R(\yseq, y) = \mathbbm{1}_{\text{Extract}(\yseq) = y}$, where Extract$(\cdot)$ pulls the solution from a \texttt{\textbackslash boxed\{\}} block within the solution. In general, $R$ could be any reward model with $y$ specifying the data required for reward computation.

The objective in \Cref{eq:rl-training} incentivizes the policy $\pith$ to maximize the expected reward while remaining close to a reference model $\piref$. This constraint is enforced via a \textit{reverse} KL divergence penalty in the objective, which is an expectation under the learned policy $\pi_{\theta}$: \looseness=-1
\begin{align}
    {\rm KL}(\pith(\cdot\mid x) \,\|\,\piref(\cdot\mid x)) = \mathbb{E}_{\yseq \sim \pith(\cdot | x)}\left[\log \frac{\pith(\yseq | x)}{\piref(\yseq | x)}\right]
\end{align}
Additionally, a control variate in the form of a \textit{baseline} $b(x,y)$ is subtracted from the reward to reduce variance during training, resulting in the \textit{advantage} $A(x,y,\yseq) = R(x,y,\yseq) - b(x,y)$\footnote{The baseline is chosen to have an expected value of $0$ under the policy and does not affect the optima.}. 

\subsection{Unbiased Gradient of a KL Estimator}
\label{sec:estimators}

In this work, 
we are interested in estimating the reverse KL divergence between the \emph{sequence-level distributions} $\pith(\yseq \mid x)$ and $\piref(\yseq \mid x)$. It is commonly estimated by decomposing into token-level estimates~\citep{sheng2025hybridflow,vonwerra2022trl,hu2024openrlhf,cui2025process}. For a generic sequence-level reverse KL divergence estimator $\widehat{\mathrm{KL}}$,
\begin{equation}\label{eq:estimator}
\begin{aligned}
D_{\mathrm{KL}}(\pith \,\|\, \piref) = \expec\!\left[\klest\right] = \expec\!\left[\sum_{t=1}^{T}\klest_{t}\right],
\end{aligned}
\end{equation}
where $\klest_t$ is the estimator defined on the \emph{token-level distributions} $\pith(\yt \mid \hist)$ and $\piref(\yt \mid \hist)$. Henceforth, any reference to an estimator implies reference to the use of the token-level version $\klest_t$. The gradient of the expectation of $\klest$, under sequences sampled from $\pith$ can then be written as
\begin{equation}
\label{eq:est_grad}
\begin{aligned}
&\d \expec\!\left[\klest\right]
=\expec\!\left[
\sum_{t=1}^{T}\d\klest_t
\right. 
\left.
+ \sum_{t=1}^{T}\klest_t\,\d\log\pith(\yseq \mid x)
\right].
\end{aligned}
\end{equation}
Such gradient estimators have been previously studied in \citet{ranganath2014black,tang2025few,zhang2025design}. See \Cref{app:est_grad} for a derivation of \Cref{eq:est_grad}. Next, we discuss two ways of using the estimator in the context of RL training of LLMs. We restrict our primary analysis in this section to the on-policy setting. 

\begin{table}[t]
\caption{\textbf{Summary of estimators considered in this study and the bias of their gradients with respect to the true gradient of sequence-level reverse KL divergence in on-policy settings.} We study 4 settings, including the commonly used \textit{$\naive$ estimator in reward} and \textit{\lowvar estimator in loss}. All configurations except using $\naive$-in-reward lead to biased gradients. In two of the cases of biased gradients, we observe training instabilities or collapses when they are used in RL fine-tuning of LLMs. $r_t = \frac{\piref(\yt \mid \hist)}{\pith(\yt \mid \hist)}$ in the expressions given.\label{tab:overview}}
\vspace{-0.25em}
\begin{center}
\begin{tabular}{ccccc}
\toprule
Estimator            &        Expression            &        Position           & Unbiased Grad. Est. (\Cref{sec:estimators}) &  Training (\Cref{sec:results}) \\
\midrule
\multirow{2}{*}{\textbf{\naive}} & \multirow{2}{*}{$-\log r_t$} & Reward  & \CheckmarkBold & Stable \\
 & & Loss & \XSolidBrush & Unstable  \\
\multirow{2}{*}{\textbf{\lowvar}} & \multirow{2}{*}{$r_t - 1 - \log r_t$} & Reward &  \XSolidBrush & Collapse \\
 & & Loss & \XSolidBrush & Stable \\
\bottomrule
\end{tabular}
\end{center}
\vspace{-1em}
\end{table}

\subsection{Placement of the KL Estimator}
\label{sec:pos_grad}

We now discuss the different ways KL estimators have been used in the RL objective -- namely, adding the estimator to the reward, and adding it directly to the loss objective.
\paragraph{Reward.} An estimator is added to the reward by applying a stop-gradient operation on the KL estimate and adding it to the token-level task score. The gradient corresponding to the KL penalty is then computed as $\expec\left[-\sum_{t=1}^{T}\klest_t \cdot \d \log \pith(\yseq \mid x)\right].$ 
\paragraph{Loss.} This refers to adding the KL estimator to the loss, as popularized by GRPO~\citep{guo2025deepseek,shao2024deepseekmath}. Automatic differentiation, which is commonly used in practice, cannot backpropagate through the sampling process used to compute the KL estimate. The gradient of the objective is then computed as $\expec\left[- \sum_{t=1}^{T} \d \klest_t\right].$

We refer the reader to Appendix~\ref{app:true_grad} and \ref{app:est_grad} for a derivation of the expressions discussed in this section. Note that in the general case, both of these terms in isolation are biased with respect to the correct gradient as stated in \Cref{eq:est_grad}. In the case of on-policy training, we can recover the correct gradient up to a normalizing factor by adding the estimator to both the reward and the loss.

\section{Inspecting KL Estimators}
\label{sec:est_fin}

In principle, if $\widehat{\rm KL}$ is an estimator for the reverse KL divergence in the sense of \cref{eq:estimator}, then it should satisfy:
\begin{equation}
\label{eq:kl_grad}
\begin{aligned}
&\d \expec[\widehat{\mathrm{KL}}] = \expec\left[
\log \frac{\pith(\yseq \mid x)}{\piref(\yseq \mid x)}
\d \log \pith(\yseq \mid x)\right],
\end{aligned}
\end{equation}
where the left side is given by \cref{eq:est_grad} and the right side is the gradient of the true sequence-level reverse KL divergence.

We now examine two commonly used unbiased KL estimators used in RL training of LLMs, namely the naïve or $\naive$ estimator, and the Schulman or \lowvar estimator~\citep{schulman2020kl}, relative to the true gradient in \Cref{eq:kl_grad}. While the K2 or squared estimator is another known KL estimator, we restrict the scope of this paper to unbiased and popularly used estimators. We refer the reader to Appendix~\ref{app:k1} and \ref{app:k3} for detailed derivations of the gradient expressions mentioned in this section. Throughout the discussion, $r_t = \frac{\piref(\yt \mid \hist)}{\pith(\yt \mid \hist)}$.

\subsection{K1 Estimator}

The $\naive$ estimator is the Monte Carlo estimate of the log-ratio of likelihoods under the current and reference policies, with samples from the current training policy. Given below are the expressions of the estimator and the gradients induced by its use in reward and loss. The gradient estimate induced when \naive ~is used in the reward is unbiased, whereas it is biased when the estimator is used in loss.
\begin{align}
    \label{eq:naive}
     \naive = \sum_{t=1}^T \naivet = \sum_{t=1}^T \log \frac{\pith(\yt \mid \hist)}{\piref(\yt \mid \hist)} = -\sum_{t=1}^T \log r_t.
\end{align}

\paragraph{Reward.} 
\begin{equation}
\label{eq:naive_grad2}
\begin{aligned}
\expec\!\left[\sum_{t=1}^{T}\naivet \d \log \pith(\yseq \mid x)\right] = \expec\!\left[
- \sum_{t=1}^T \log r_t \cdot
\d\log \pith(\yseq \mid x)\right].
\end{aligned}
\end{equation}

\paragraph{Loss.} 
\begin{equation}
\label{eq:expec_zero}
\begin{aligned}
\expec\!\left[\sum_{t=1}^{T}\d\naivet\right] = \expec\!\left[
\d \log \frac{\pith(\yseq \mid x)}{\piref(\yseq \mid x)}
\right] = 0.
\end{aligned}
\end{equation}
$\naive$ is a special case where the gradient of the expectation of the estimator obtained from adding KL estimator in the reward results in an unbiased estimator~\citep{tang2025few}. This is because the first term of equation \Cref{eq:est_grad} is zero in expectation for the $\naive$ estimator as shown in \Cref{eq:expec_zero}.

\subsection{K3 estimator}
The $\lowvar$  has a lower variance and thus is often preferred over $\naive$ in practice. $\lowvar$ induces biased gradient estimates both when used in reward or loss. The estimator and gradient expressions are as given below. 
{\
\begin{align}
    \label{eq:lowvar}
    \lowvar = \sum_{t=1}^T \lowvart = \sum_{t=1}^T \Bigg[\frac{\piref(\yt \mid \hist)}{\pith(\yt \mid \hist)} - 1 - \log\frac{\piref(\yt \mid \hist)}{\pith(\yt \mid \hist)}\Bigg] = \sum_{t=1}^T \left[r_t - 1 - \log r_t\right]. 
\end{align}
}

\paragraph{Reward.} 
\begin{equation}
\begin{aligned}
\d \expec\!\left[\sum_{t=1}^{T}\lowvart\d\log \pith(\yseq \mid x) \right] = \expec\!\Bigg[
\sum_{t=1}^{T}\left(r_t - \log r_t\right)
\cdot\d \log\pith(\yseq \mid x)\Bigg].
\label{eq:k3_in_reward}
\end{aligned}
\end{equation}


\paragraph{Loss.} 
\begin{equation}
\begin{aligned}
\expec\!\left[\sum_{t=1}^{T}\d\lowvart\right] = \expec\left[-
\sum_{t=1}^{T}
r_t \cdot\d \log \pith(\yt \mid \hist)
\right].
\label{eq:k3_in_loss}
\end{aligned}
\end{equation}

In particular, the gradient of the expectation of the KL estimate when $\lowvar$ is used in the loss is a biased estimate of the true reverse KL gradient shown in \Cref{eq:est_grad}. This is the version used in the implementations of some of the most popular RL algorithms for LLMs, such as GRPO~\citep{shao2024deepseekmath,guo2025deepseek}.

\paragraph{Off-policy training.} In the off-policy setting, none of the KL configurations we discuss are unbiased due to the difference in the sampling and training policy. \citet{zhang2025design} note that an unbiased gradient estimate of the \textit{token-level} reverse KL gradient can be computed as the token-level importance sampling ratio $\omega_t$ weighted token-level KL estimate added to the loss directly (\ie, $\omega_t \lowvart$ or $\omega_t \naivet$). However, we focus on \emph{sequence-level} reverse KL objective in this work. The sequence-level reverse KL objective implementations resulting in unbiased reverse-KL estimates are susceptible to high variance from the sequence-level importance sampling ratio $\omega_s$ for long sequences. In practice, the sequence-level KL objective is weighed by the token-level importance sampling ratios when KL is used in the reward, and the importance ratio is skipped entirely when used in the reward (as in GRPO). Thus, for the off-policy setting, we primarily study the empirical performance of the KL configurations.


\subsection{Parametric Autoregressive Model: An Illustrative Example}
\label{sec:toy}

We first study the bias in the gradients of the KL estimators in a minimal parametric autoregressive model. We define reference models $A$ and $B$ over binary sequences, each factorizing into Bernoulli conditionals over each token in the sequence (conditioned on the previous tokens).
\begin{equation}
\label{eq:AB_defs}
\begin{aligned}
A_\theta(Y)
&= \prod_{t=1}^{T} (p_t^A)^{y_t}\,(1-p_t^A)^{1-y_t},
p_t^A = \sigma\!\big(a + b\,c_{t-1}\big), \quad c_{t-1}
= \sum_{k=1}^{t-1} y_k .\\
B_\phi(Y)
&= \prod_{t=1}^{T} (p_t^B)^{y_t}\,(1-p_t^B)^{1-y_t}, p_t^B
= \sigma\!\big(\tilde a + \tilde b\,c_{t-1}\big), \quad c_{t-1}
= \sum_{k=1}^{t-1} y_k .
\end{aligned}
\end{equation}
Here, $Y$ is a binary sequence, $a, b, \tilde a$ and $\tilde b$ are the parameters of distributions $A$ and $B$, $c_0 = 0$, $\yt \in\{0,1\}$ and $\sigma$ represents the sigmoid function. The reverse KL $D_{\mathrm{KL}}(A\,\|\,B)$ and its gradient with respect to $a$ and $b$ admit closed-form expressions (see \cref{app:toy_grad}).

\begin{figure}[t]
    \centering
    \includegraphics[width=\linewidth]{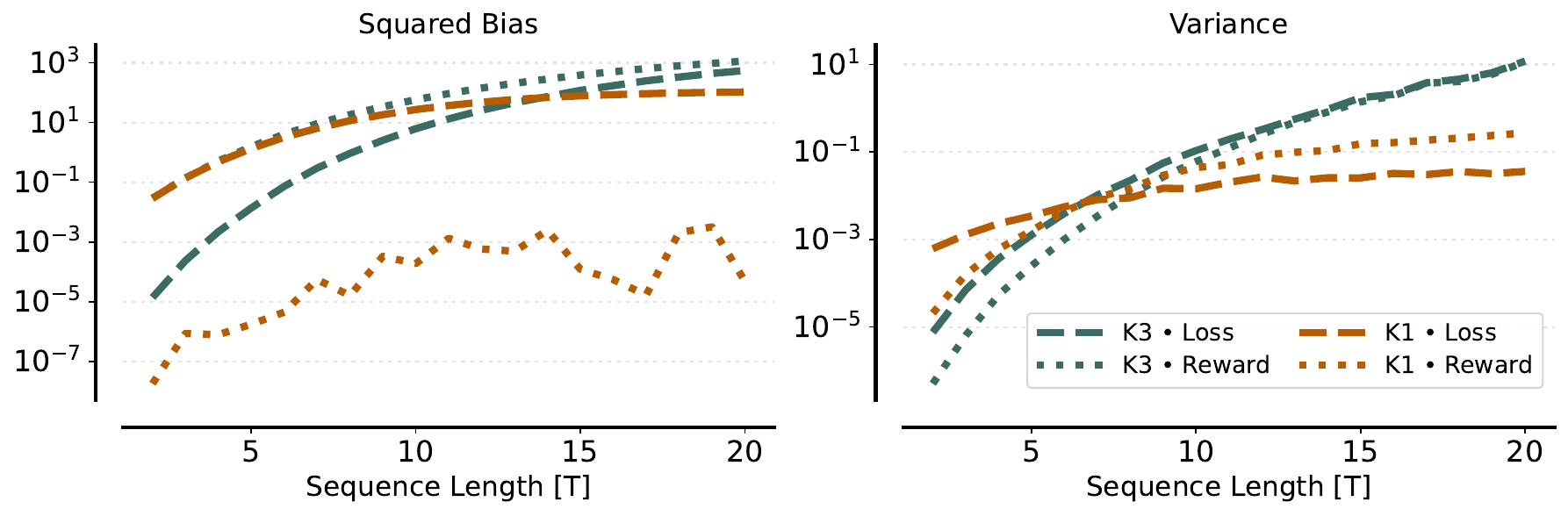}
    \vspace{-2em}
    \caption{\textbf{The bias and variance of expected gradients with respect to the parameters of $A$, in different configurations (logarithmic scale).} While all estimators are unbiased, the expected gradients are unbiased only in the case of $\naive$ estimator when used in reward. \lowvar estimator when used in reward exhibits the highest gradient  bias. While $\naive$ estimator in loss has relatively lower variance in the estimated gradient, it also suffers from high bias.}
    \label{fig:toy}
    \vspace{-1.5em}
\end{figure}

We compute the KL divergence with different estimators and their gradients when used as reward and in the loss as discussed in~\Cref{sec:estimators}, using $200$ trials each with $N=1000$ sequences of lengths $T$ sampled from $A$. 
We illustrate the bias and variance of the gradient estimates of the different configurations in \cref{fig:toy}. Note the logarithmic scale of the plots. We observe that the  bias of the gradient of the $\naive$ estimator added to the reward remains low. On the other hand, the gradients associated with the $\lowvar$ estimator in the loss and the reward both show high bias and variance. These results validate the conclusions from \cref{sec:estimators}. 

\section{RL Fine-tuning of LLMs}
\label{sec:results}
\vspace{-1em}

We complement our analysis in \cref{sec:est_fin} with an empirical study on the effect of design choices in KL regularization on training LLMs with RL. We use the initial base model as the reference model $\piref$ in all experiments. We conduct the following investigations:
\begin{tcolorbox}[colback=orange!10,
leftrule=0.5mm,top=1mm,bottom=1mm,boxrule=0.6pt]
\begin{itemize}[left=-0.3cm,nosep]
    \item (\cref{sec:results_onpolicy}) How do the various KL configurations discussed in \cref{sec:est_fin} affect the downstream performance of LLMs trained using on-policy RL in both in- and out-of-domain tasks?
    \item (\cref{sec:offpolicy_results}) How do different KL configurations affect the downstream performance of models when used in an asynchronous RL (i.e. off-policy) setup. 
\end{itemize}
\end{tcolorbox}

 Motivated by our observations in this section, we list the recommended settings to induce the unbiased KL gradient estimate for commonly used libraries in \Cref{tab:lib_settings}.

\subsection{Effect of KL Configurations in On-policy RL}
\label{sec:results_onpolicy}
\vspace{-0.5em}

\paragraph{Experimental Setup.} We (on-policy) RL fine-tune \texttt{Qwen2.5-7B} and \texttt{Llama-3.1-8B-Instruct} models on the training subset of Hendrycks MATH~\citep{hendrycks2021measuring} (henceforth referred to as MATH) with different KL configurations, across different values of the $\beta$ coefficient. We evaluate the trained models on 2 different in-domain tasks, namely MATH500~\citep{lightman2023let} and MATH$^2$~\citep{shah2024ai} and 3 out-of-domain tasks -- MMLU college physics, college chemistry  and college biology subsets \citep{hendrycks2021measuring}. To demonstrate training behavior, we report the train-time Pass@1 score of the models on an in-distribution but held-out test set of the task and report mean@32 accuracy across 3 seeds on the final evaluations. Models are evaluated at the same sampling temperature they are trained at. More details about the experimental setup are discussed in \Cref{app:setup}. We also validate our observations on an additional experimental setup involving \texttt{Qwen3-4B-Instruct-2507}~\citep{yang2025qwen3} and CodeIO~\citep{stojanovski2025reasoning} in ~\Cref{app:codeio}.

In order to restrict our study on the effect of the KL estimators, we opt to use REINFORCE leave-one-out advantage baseline~\citep[RLOO]{ahmadian2024back}. We use a batch size of 256, maximum response length 1024 tokens, a learning rate of $10^{-6}$ and sample $K=5$ responses per prompt at a temperature of 1.0. We discuss our main observations below.


\begin{tcolorbox}[colback=orange!10,
leftrule=0.5mm,top=1mm,bottom=1mm,boxrule=0.6pt]
\textbf{Observation 1:} \emph{Adding $\naive$ estimator to the loss can lead to training instabilities.}
\end{tcolorbox}

\setlength{\intextsep}{0pt}
\setlength{\columnsep}{10pt}
\begin{wrapfigure}{r}{0.6\linewidth}

    \centering
    \includegraphics[width=0.49\linewidth]{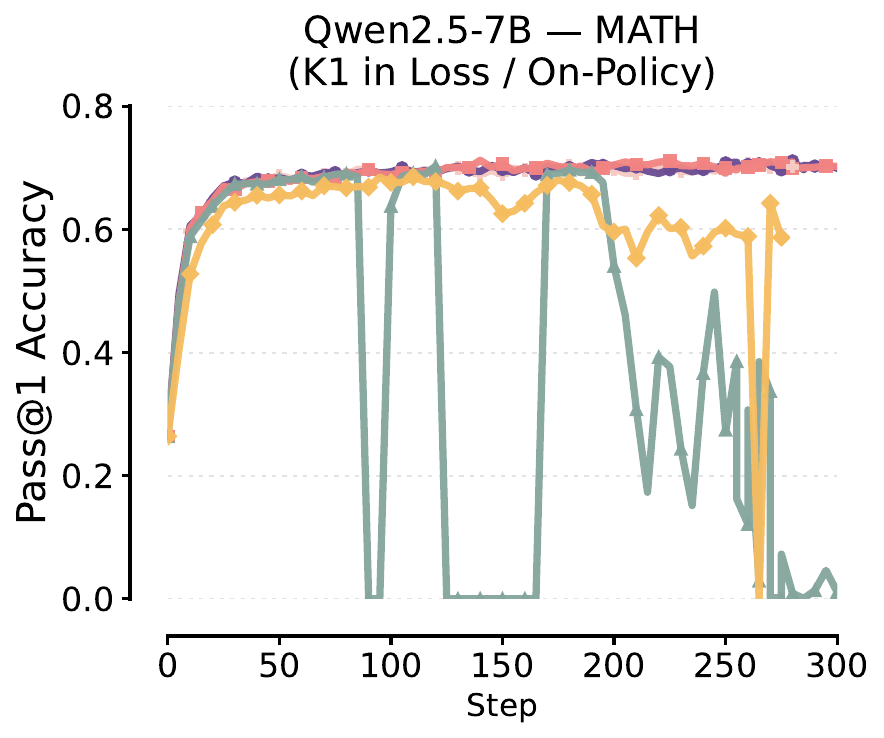}\hfill
    \includegraphics[width=0.49\linewidth]{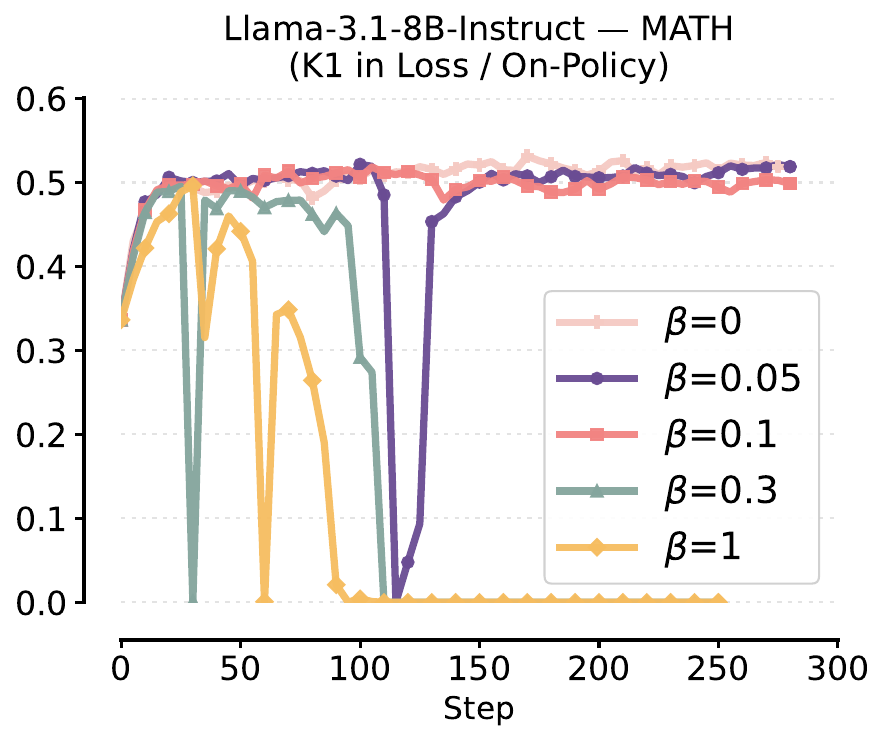}
    \vspace{-0.5cm}
    \caption{\textbf{Training Instabilities when using $\naive$-in-loss.} Pass@1  performance for \textbf{[Left]}  training \texttt{Qwen2.5-7B} with $\naivet$ leads to training instabilities for $\beta = 0.1$ and $1$. \textbf{[Right]} Training \texttt{Llama-3.1-8B-Instruct} with $\naivet$ in loss leads to instabilities for all $\beta$ except 0.1.}
    \label{fig:naive_kl_loss}
\end{wrapfigure}

As shown in \Cref{eq:expec_zero}, adding $\naivet$ to the loss results in a biased estimate of the reverse KL gradient, since the term is zero in expectation. Intuitively, RL fine-tuning with $\naivet$ with any coefficient $\beta$ should perform similar to RL fine-tuning without any KL penalty ($\beta=0$). To verify this empirically, we fine-tune \texttt{Qwen2.5-7B}~\citep{Yang2024Qwen25TR} and \texttt{Llama-3.1-8B-Instruct}~\citep{touvron2023llama} with $\beta = 0.05, 0.1, 0.3$ and $1$ and compare them against RL fine-tuning with $\beta=0$. \Cref{fig:naive_kl_loss} shows Pass@1 performance of models on MATH test set over the course of training.

We observe training instabilities across both \texttt{Qwen2.5-7B} and \texttt{Llama-3.1-8B-Insturct} models, and different values of $\beta$ as shown in  \Cref{fig:naive_kl_loss}. A potential explanation is that the term $\sum_{t=1}^{T}\d\log\pith(\yt \mid \hist)$, despite having an expectation of 0, adds variance to the optimization, leading to instabilities. 
In cases where the training is stable, i.e., \texttt{Qwen2.5-7B} trained with $\beta = 0.05$ and $0.1$, the performance is similar to training without any KL, as expected. \texttt{Qwen2.5-7B} seems to be more robust to variance as compared to \texttt{Llama-3.1-8B-Instruct} models. 

\begin{figure}[b]
    \centering
    \includegraphics[width=0.245\linewidth]{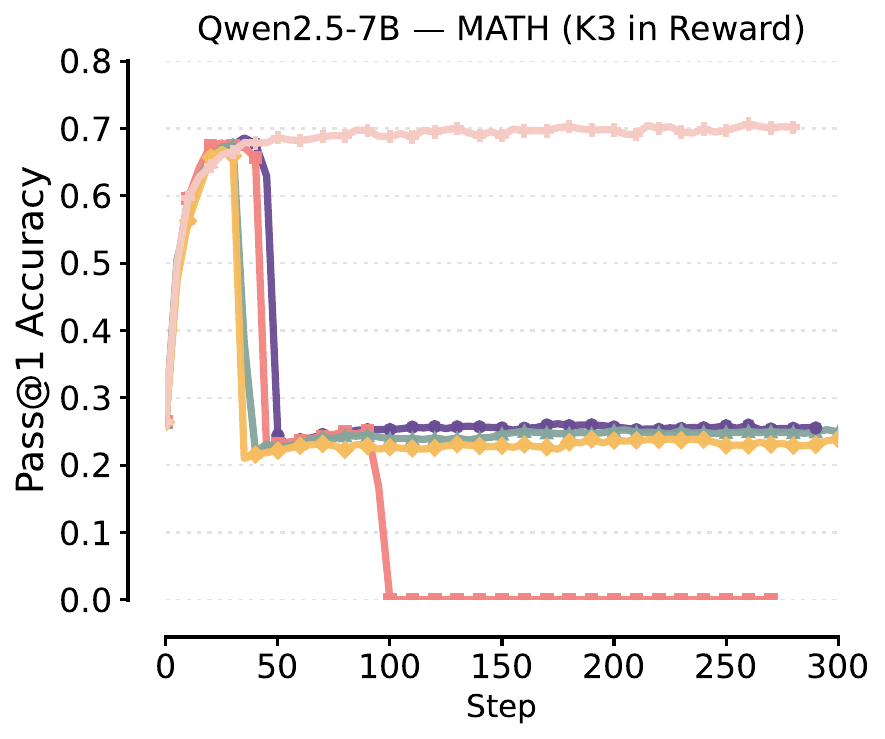}
    \includegraphics[width=0.245\linewidth]{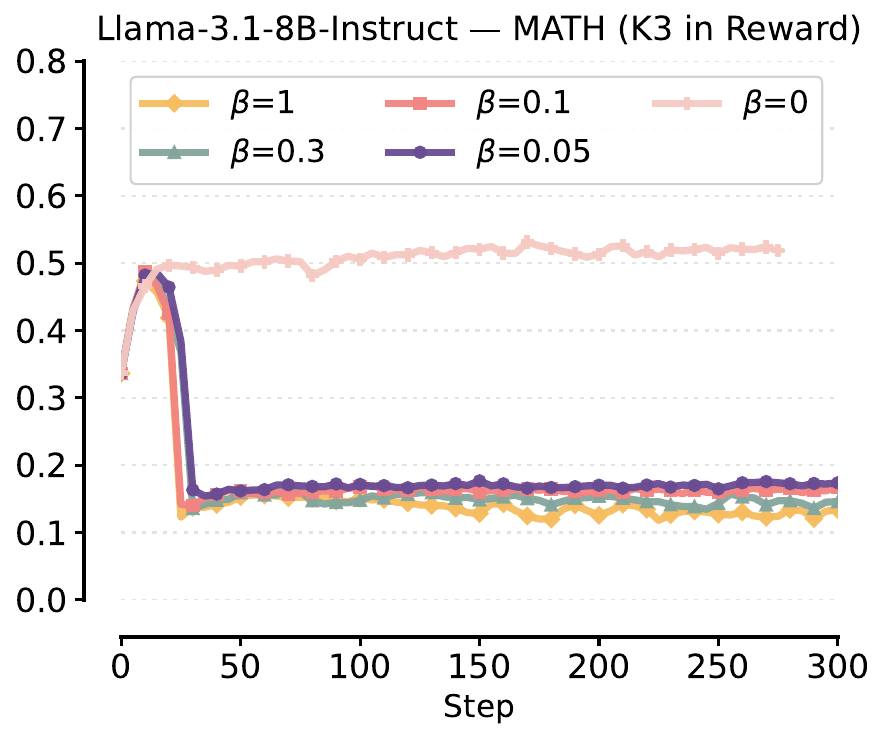}
    \includegraphics[width=0.245\linewidth]
    {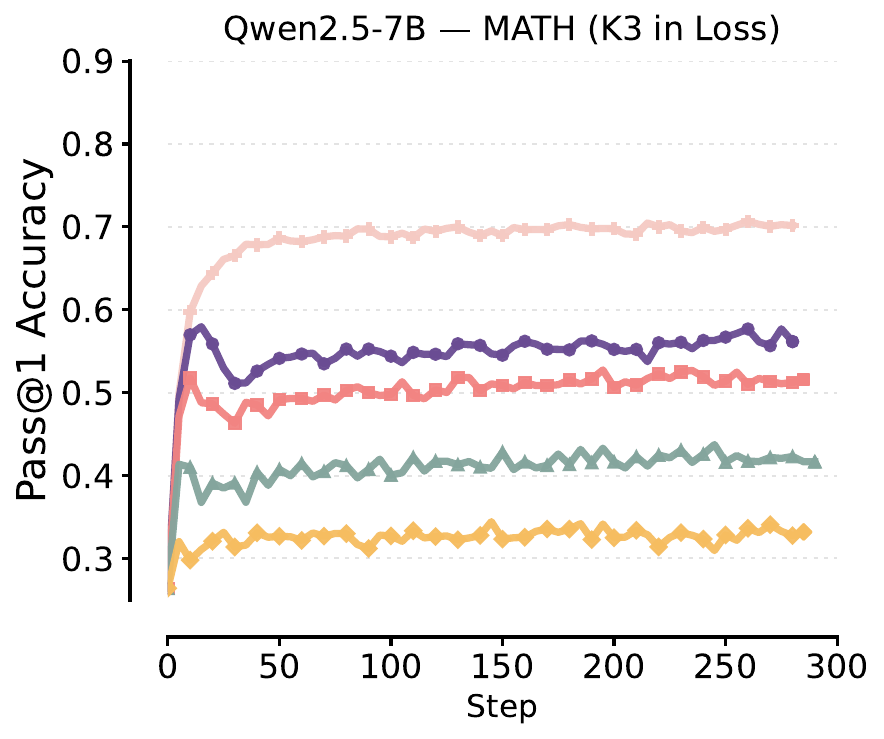}
    \includegraphics[width=0.245\linewidth]
    {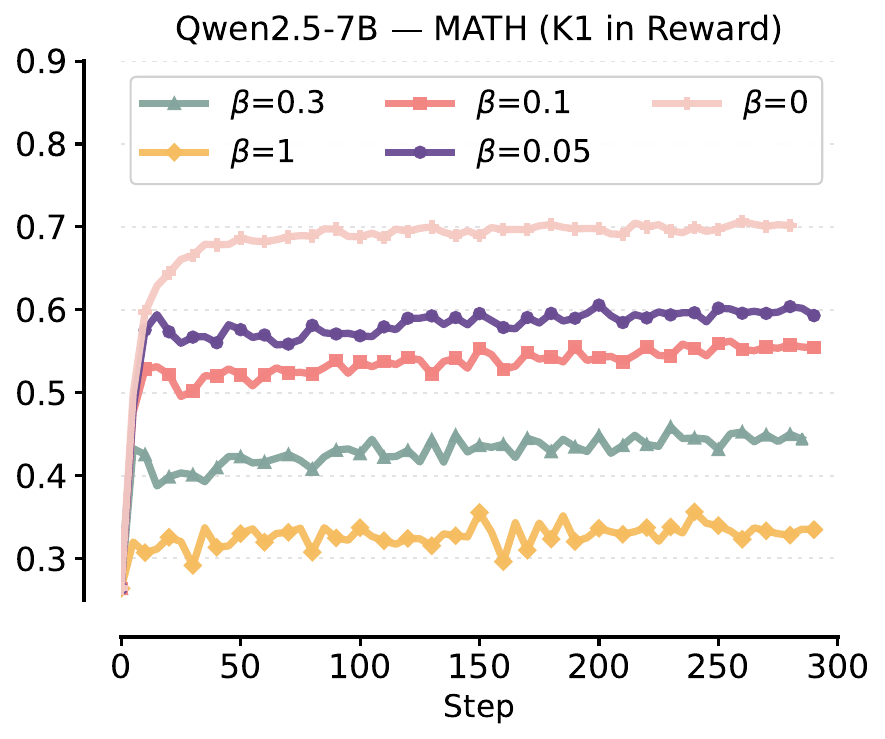}
    \vspace{-2em}
    \caption{\textbf{Collapse in the case of adding \lowvar to the reward.} \textbf{First:} Pass@1 performance for \texttt{Qwen2.5-7B} trained on MATH train dataset. \textbf{Second:} \texttt{Llama-3.1-8B-Instruct} trained on MATH train dataset with \lowvar-in-reward. The collapse may be attributed to high bias and variance of the configuration. \textbf{Pass@1 on MATH test set with \lowvar-in-loss (biased; third) and \naive-in-reward (unbiased; fourth).} Although biased with respect to reverse KL, \lowvar-in-loss yields stable training. In both cases, lower $\beta$ values lead to higher performance.}
    \label{fig3:k3_in_reward_loss}
    \vspace{-1em}
\end{figure}

\begin{tcolorbox}[colback=orange!10,
leftrule=0.5mm,top=1mm,bottom=1mm,boxrule=0.6pt]
\textbf{Observation 2:} \emph{Adding \lowvar estimator to the reward leads to training collapse.}
\end{tcolorbox}
Another case of biased gradient estimate is \lowvar used in the reward. From \Cref{eq:k3_in_reward}, we observe that adding token-level \lowvar to rewards leads to a bias term of
\begin{align}
\expec\left[\sum_{t=1}^{T}\frac{\piref(\yt \mid \hist)}{\pith(\yt \mid \hist)}\d\log\pith(\yseq \mid x)\right].
\end{align}
This bias is illustrated in~\Cref{fig:toy} for the parametric autoregressive model~\Cref{sec:toy}. To validate this in the case of LLMs, we RL fine-tune \texttt{Qwen2.5-7B} and \texttt{Llama-3.1-8B-Instruct} with $\beta = 0.05, 0.1, 0.3$ and $1$. \Cref{fig3:k3_in_reward_loss} (\textbf{first} and \textbf{second}) shows that this biased gradient estimate results in unpredictable behavior leading to complete or partial collapse for all $\beta$. 


\begin{tcolorbox}[colback=orange!10,
leftrule=0.5mm,top=1mm,bottom=1mm,boxrule=0.6pt]
\textbf{Observation 3:} \emph{Unbiased gradient estimator results in better out-of-distribution performance as compared to biased estimators with stable training behaviors.} 
\end{tcolorbox}
\looseness=-1
The final setting leading to a biased expected gradient of the sequence-level reverse KL is when \lowvar is used in the loss \Cref{eq:k3_in_loss}. Surprisingly, despite the bias, using \lowvar in the loss exhibits stable training of the policy. \Cref{fig3:k3_in_reward_loss} (\textbf{third}) reports the in-distribution performance of \texttt{Qwen2.5-7B} models (evaluated on MATH test dataset) for different $\beta$ during training. This may be explained by the observation that the gradient estimate \Cref{eq:k3_in_loss} in this case is a sum of unbiased gradient estimates of the forward KL divergences computed at the token level, making this configuration equivalent to minimizing the forward KL divergence for each token (see \Cref{app:k3}). Note that using \lowvar-in-loss is a popular configuration used with policy optimization algorithms such as GRPO~\citep{shao2024deepseekmath,guo2025deepseek}.

\begin{figure}
    \centering
    \includegraphics[width=1\linewidth]{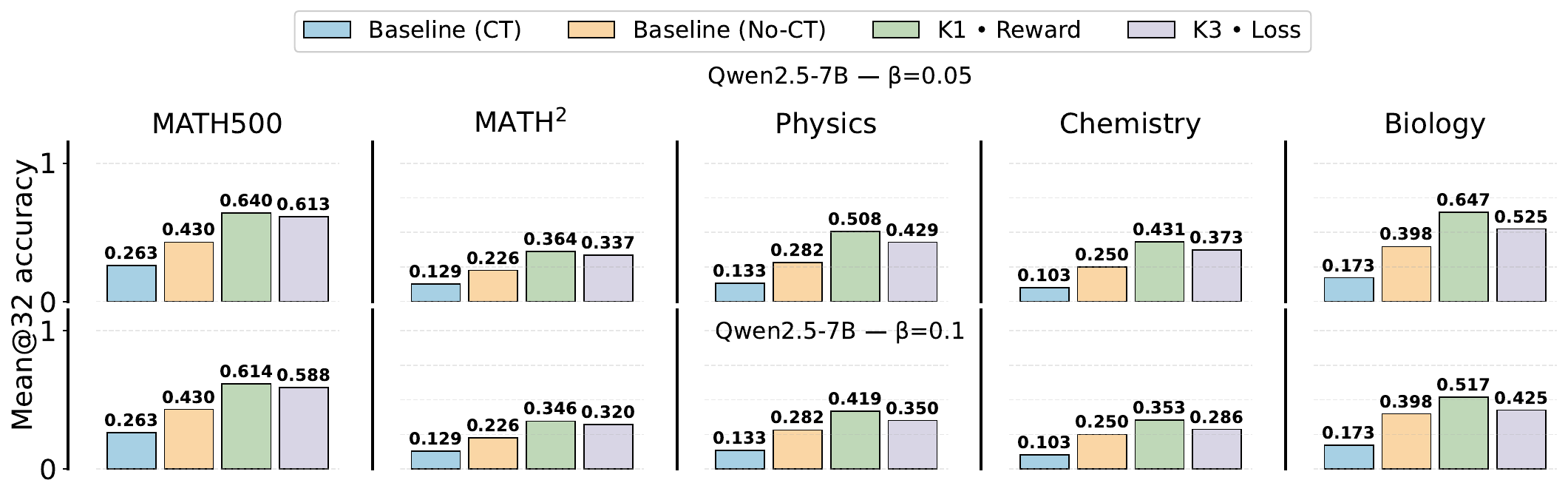}
    \vspace*{-2em}
    \caption{\textbf{Comparison of \texttt{Qwen2.5-7B} trained with two stable estimator configurations - $\naive$-in-reward and \lowvar-in-loss.} Baseline (CT) refers to the performance of base \texttt{Qwen2.5-7B} when prompted with a chat template. Baseline (No-CT) represents the performance when it is prompted without a chat template. $\naive$-in-reward (unbiased gradient)  performs the best on both in-domain and out-of-domain tasks. Increasing $\beta$
    deteriorates performance.}
    \label{fig:ood}
    \vspace{-2em}
\end{figure}

$\naive$-in-reward, resulting in an unbiased estimate of the true gradient (\Cref{eq:naive_grad2}), demonstrates stable training as shown in \Cref{fig3:k3_in_reward_loss} (\textbf{fourth}). We compare the downstream performance of the models trained with \lowvar-in-loss and $\naive$-in-reward. We train \texttt{Qwen2.5-7B} using these two estimator configurations for 250 steps on MATH train data, and compare the performance of the two configurations on a wide range of evaluation tasks as shown in \Cref{fig:ood}.
Similar results for \texttt{Qwen2.5-7B} with $\beta = 0.3$ and $\beta = 1$ can be found in \cref{app:add_eval}. 

We observe that using $\naive$-in-reward (\ie \space unbiased gradient estimate) outperforms using \lowvar-in-loss (\ie \space biased gradient estimate). While the performance gains are consistent across all tasks and both models, we observe that the gains are more pronounced in out-of-domain tasks for \texttt{Qwen2.5-7B}, with an average relative improvement of 19.06\% across MMLU college-physics, college-chemistry, and college-biology, as compared to an average relative improvement of only 6.21\% on in-domain tasks, for $\beta=0.05$. On the other hand, the gains are more pronounced in-domain (average relative improvement of 15.94\%) than out-of-domain (average relative improvement of only 3.65\%) for \texttt{Llama-3.1-8B-Instruct}.

We investigate the effect of using the KL term in both the reward and the loss (resulting in unbiased gradient estimates for both \naive and \lowvar, \Cref{sec:pos_grad}) in \Cref{sec:correct_results}, and we perform further analysis for \lowvar-in-loss and \naive-in-reward in \Cref{app:analysis}.

\subsection{Effect of KL Configurations in Asynchronous RL}
\label{sec:offpolicy_results}

\setlength{\intextsep}{7pt}
\setlength{\columnsep}{10pt}
\begin{figure}[]
    \vspace{-13pt}
    \hspace{-6pt}
    \centering
    \includegraphics[width=0.33\linewidth]{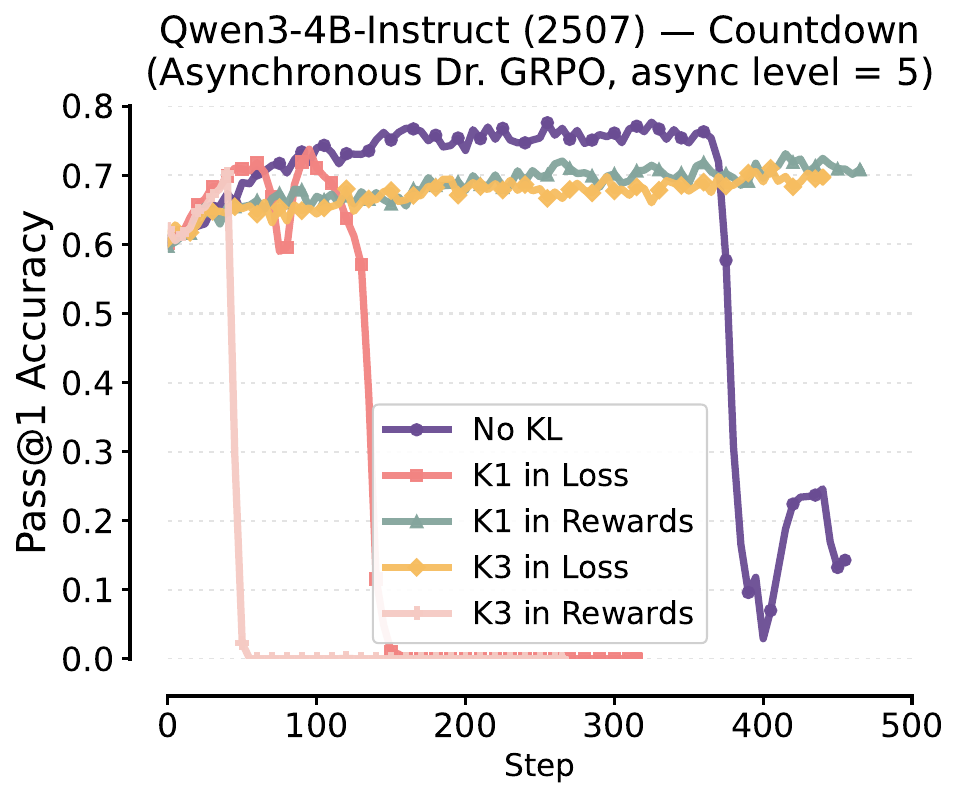}
    \includegraphics[width=0.33\linewidth]{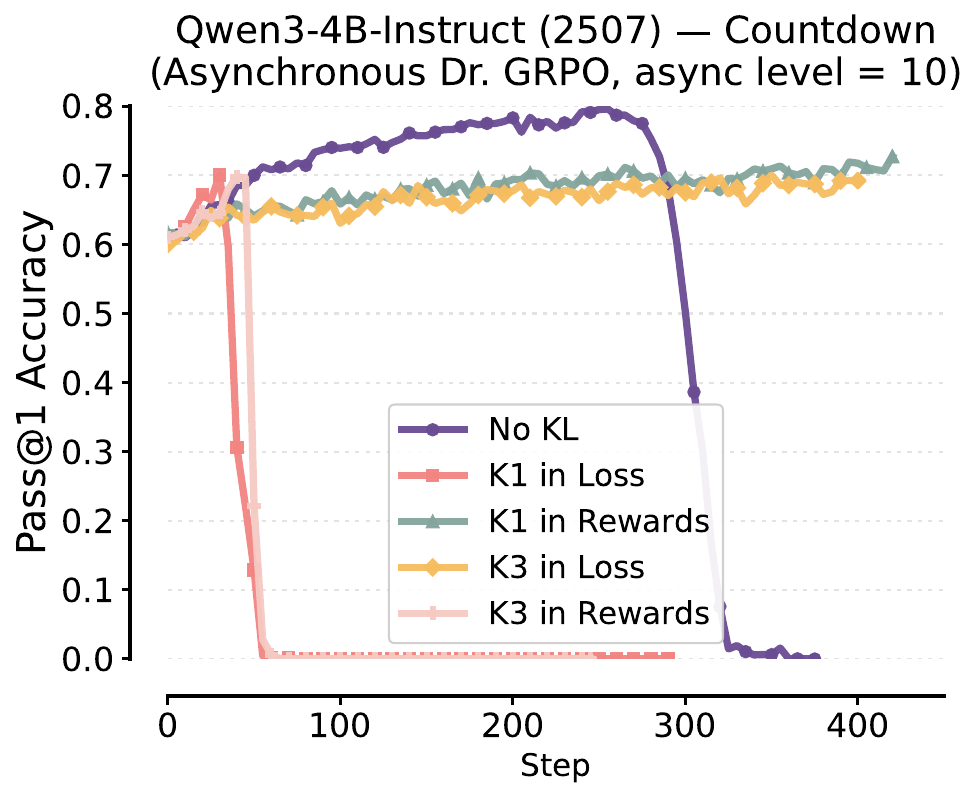}
    \includegraphics[width=0.33\linewidth]{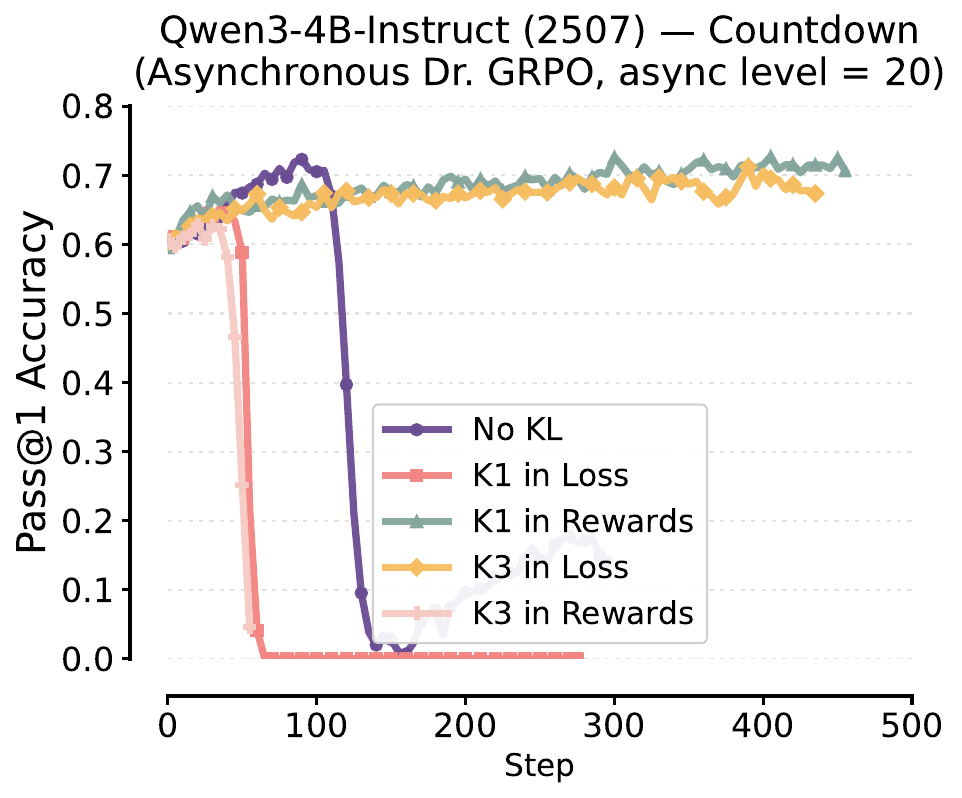}
    \vspace{-1em}
    \caption{\textbf{Comparison of different KL configurations in asynchronous RL training of \texttt{Qwen3-4B-Instruct-2507} on Countdown (async levels = 5, 10, 20).} Using \naive-in-reward and \lowvar-in-loss show stable training as opposed to using no KL regularization, \naive-in-loss and \lowvar-in-reward.}
    \label{fig:async_train}
\end{figure}

While all four KL configurations discussed result in biased sequence-level KL gradient updates in off-policy settings (as noted in \Cref{sec:est_fin}), we study the performance of the four configurations empirically. We use an asynchronous RL setup~\citep{noukhovitch2024asynchronous,bartoldson2025trajectory} which is commonly used for large-scale RL training runs in order to reduce latency. The asynchronous policy updates give rise to off-policyness.

\begin{wrapfigure}[18]{r}{0.5\textwidth}
    \vspace{-0.8\baselineskip}
    \centering

    \includegraphics[
        width=0.8\linewidth
    ]{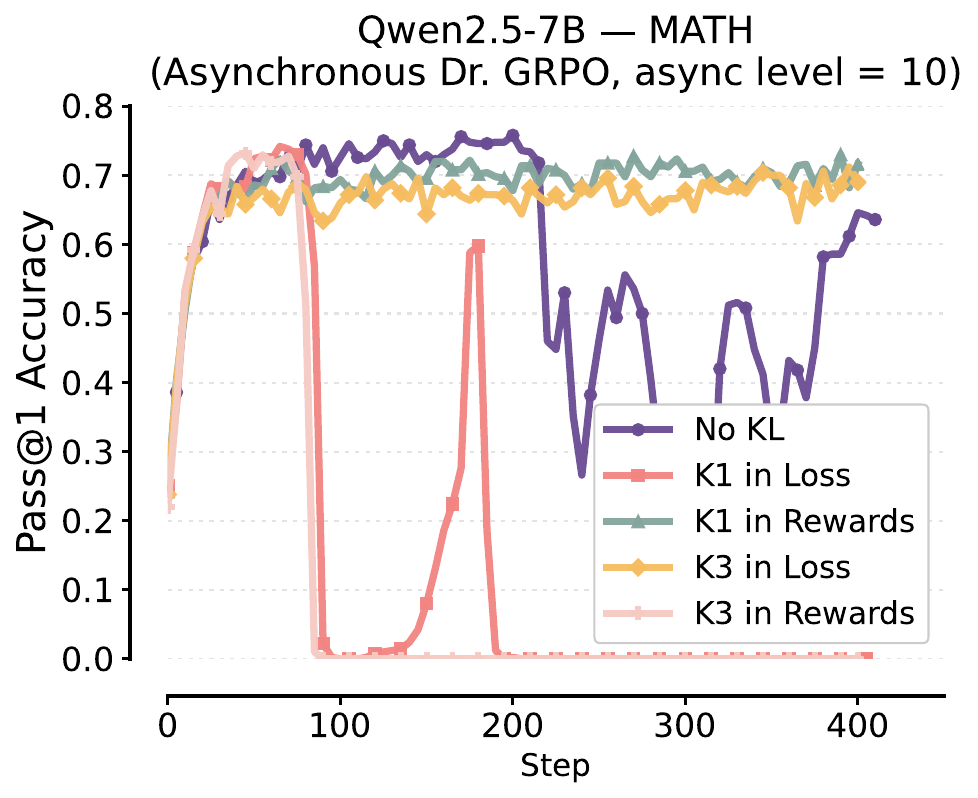}

    \captionsetup{
        font=small,
        justification=raggedright,
        singlelinecheck=false
    }
    \vspace{-3mm}
    \caption{
        \textbf{Comparison of different KL configurations in asynchronous RL training of \texttt{Qwen3-2.5-7B} on Hendrycks MATH (async level = 10).} Staying consistent with the observations in async training of \texttt{Qwen3-4B-Instruct-2507} + Countdown, using \naive-in-reward and \lowvar-in-loss show stable training as opposed to using no KL regularization, \naive-in-loss and \lowvar-in-reward.
    }
    \label{fig:math_async_train}

\end{wrapfigure}

\paragraph{Experimental Setup.} We train \texttt{Qwen3-4B-Instruct-2507}~\citep{yang2025qwen3} on Countdown~\citep{stojanovski2025reasoning}, with Dr. GRPO~\citep{liu2025understanding} at a high asynchrony levels of 5, 10 and 20. We use a learning rate of $10^{-6}$, sample 16 rollouts per prompt at a temperature of 1.0, use a training batch size of 512, maximum response length of 4096, and train for up to 350  steps, with $\beta = 0.005$. The rest of the experimental setup remains the same as in the on-policy experiments. \Cref{fig:async_train} shows the test performance across different KL configurations for all async levels, during the course of training. The plots show the performance on a held-out set of countdown.

\begin{figure}[t]
    \centering
    \includegraphics[width=0.85\linewidth]{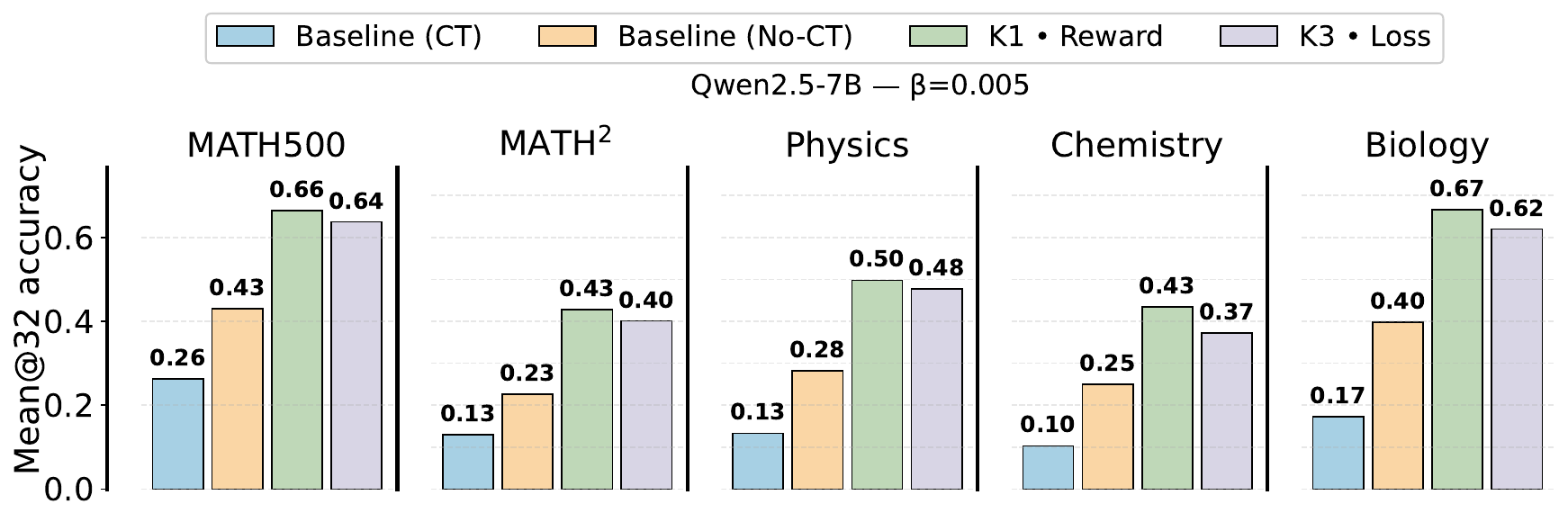}
    \vspace{-1em}
    \caption{\textbf{Comparison of Qwen2.5-7B trained on MATH dataset in asynchronous setting with the two stable configurations - $\naive$-in-reward and \lowvar-in-loss.} Consistent with the observations in synchronous settings, \naive-in-reward outperforms \lowvar-in-loss across all considered evaluation tasks.}
    \label{fig:async_eval}
    \vspace{-1em}
\end{figure}

\vspace{5mm}
Further, we also train \texttt{Qwen2.5-7B} on MATH with the same experimental setup and an async level of 10. \Cref{fig:math_async_train} shows the performance on the test set of Hendrycks MATH over the course of the training. \Cref{fig:async_eval} shows the performance of the final checkpoints fine-tuned with $\naive$-in-reward, and \lowvar-in-loss (the two stable settings) on other in-domain and out-of-domain evaluation sets. Note that the performance gap between no KL training and KL regularized training can be closed by using regular reference policy resets as noted in~\cite{bartoldson2025trajectory}.


\begin{tcolorbox}[colback=orange!10,
leftrule=0.5mm,top=1mm,bottom=1mm,boxrule=0.6pt,breakable]
\textbf{Our observations are twofold:}
\begin{itemize}[left=0cm,nosep]
    \item In highly asynchronous settings, $\naive$-in-reward and \lowvar-in-loss, both help stabilize the training in cases where it is otherwise unstable in the absence of any KL regularization.
    \item Staying consistent with observations in the case of on-policy RL, the use of $\naive$-in-loss and \lowvar-in-reward show training instabilities whereas the use of $\naive$-in-reward and \lowvar-in-loss show stable training. In evaluation, $\naive$-in-reward outperforms \lowvar-in-loss.
\end{itemize}
\end{tcolorbox}


\section{Conclusion}
\label{sec:conclusion}

\looseness=-1
We conduct a study of how different KL estimators, and their placement within the RL objective, affect the stability and performance of RL fine-tuning of LLMs. We consistently find that implementations with biased reverse KL divergence gradient estimates perform unpredictably: at worst leading to training collapses and at best still underperforming implementations with unbiased gradient estimates. While the $\lowvar$ estimator in the loss, commonly used in GRPO, remains generally stable, it consistently underperforms the naïve $\naive$ in reward configuration. These results all suggest that unbiased gradient configurations should serve as the default for stable and generalizable RL post-training. Updating parameters with a step that is not the gradient of the desired objective will produce undesired optima; furthermore, making steps along a non-conservative vector field is a recipe for instability, as stable behavior near an optimum is not guaranteed. We hope that our systematic study will help clarify these issues for the community.



\section*{Ethics Statement}
Our work focuses on studying the effect of various design choices for KL regularization in training language models with RL on the performance on downstream tasks. While we foresee no immediate negative societal consequences of our work, we hope that future researchers who build upon it will bear in mind the potential of LLMs -- and in particular of human-like reasoning in LLMs -- to be used both for good and for harm.

\section*{Acknowledgments}
The authors thank Anirudh Goyal, Michael Noukhovitch and Emiliano Penaloza for interesting discussions and helpful comments. The research was enabled in part by computational resources provided by the Digital Research
Alliance of Canada (\url{https://alliancecan.ca}), Mila (\url{https://mila.quebec}), NVIDIA, and the National Energy Research Scientific Computing Center, a DOE Office of Science User Facility supported by the Office of Science of the U.S. Department of Energy under Contract No. DE-AC02-05CH11231 using NERSC award ASCR-ERCAP0032802 and NERSC award ASCR-ERCAP0032812.

AC and YB acknowledge funding from National Sciences and Engineering Council of Canada (NSERC) and the Canadian Institute for Advanced Research (CIFAR). AC acknowledges funding from the AI for Math fund (\url{https://www.renaissancephilanthropy.org/ai-for-math-fund}) and Google, supporting the VS and JO graduate programs. GB acknowledges funding from NSERC and CIFAR. ESW acknowledges support from the CIFAR Learning in Machines and Brains program. VS was supported by the UNIQUE excellence scholarship (\url{https://www.unique.quebec/}) for a partial duration of the work. MJ is supported by a FRQNT Doctoral Fellowship (\url{https://doi.org/10.69777/366694}). SM acknowledges funding from FRQNT Doctoral Fellowship (\url{https://doi.org/10.69777/372208}).

This manuscript has been authored by Lawrence Livermore National Security, LLC under Contract No. DE-AC52-07NA27344 with the U.S. Department of Energy. This material is based upon work supported by the Department of Energy (LLNL-LDRD Program under Project No. 24-ERD-058, LLNL-CONF-2014473), Office of Science, Office of Advance Scientific Computing Research. The United States Government retains, and the publisher, by accepting the article for publication, acknowledges that the United States Government retains a non-exclusive, paid-up, irrevocable, worldwide license to publish or reproduce the published form of this manuscript, or allow others to do so, for United States Government purposes.

\bibliography{colm2026_conference}
\bibliographystyle{colm2026_conference}

\clearpage
\appendix
\etocdepthtag.toc{appendix}
\etocsettocstyle{\section*{Appendix Contents}}{} 
\setcounter{tocdepth}{3}
\etocsettagdepth{mtoc}{none}           
\etocsettagdepth{appendix}{subsection}  
\tableofcontents                        

\clearpage
\section{Related Work}
\label{sec:related_work}

KL divergence-based regularization in RL comes from the literature on KL control within stochastic optimal control~\citep{todorov2006linearly,ziebart2008maximum,kappen2012optimal,rawlik2012stochastic}. The KL regularization is typically applied as a penalty term in the reward function in KL control. In the context of sequence models, \citet{jaques2017sequence} first proposed using KL regularization to maintain information learned from data in pretraining, when training the model with RL with task-specific rewards. \citet{ziegler2019fine} used this idea in the context of training language models with reward models learned from human preference data. In the context of reinforcement learning with learned reward models, KL regularization can also mitigate issues of reward-overoptimization~\citep{gao2023scaling}, where the language model being trained can learn to exploit spurious correlations in the reward model and suffer catastrophic forgetting of the natural language abilities. KL regularization remains a key ingredient in the reinforcement learning from human feedback (RLHF) pipeline~\citep{ouyang2022training,touvron2023llama}.

For language models, computing the KL divergence is intractable. In practice, sample-based Monte-Carlo estimators are used to approximate the KL divergence to be added in the reward~\citep{schulman2020kl}. Early work on RLHF~\citep{stiennon2020learning,ouyang2022training} for language models used the K1 estimator, whereas recently \citet{shao2024deepseekmath} proposed using the K3 estimator, and \citet{amini2025better} proposed a Rao-Blackwellized estimator. \citet{shao2024deepseekmath} also proposed incorporating the KL regularization term in the loss as opposed to the traditional location as part of the reward. Early work on RLHF~\citep{ziegler2019fine,stiennon2020learning} studied the effect of the strength of the KL regularization but there is no systematic study on the downstream effect of various estimators. 

In the context of RL with verifiable rewards for LLM reasoning, the role of KL has received renewed interest. \citet{guo2025deepseek} follow the GRPO recipe and propose using the K3 estimator in the loss. Some work, \eg, \citet{yu2025dapo} argue against using any KL regularization at all to allow the model to explore significantly and reduce memory usage. \citet{gx2025kl} re-examine the traditional intuition about the effects of forward and reverse KL in the context of LLM training and identify that the mode coverage instead depends primarily on factors such as regularization strength and relative scales between the rewards and the reference probabilities. \citet{wu2024rethinking} also make the same observation in the case of on-policy distillation. \citet{zhang2025design} provide a unified formulation for regularized policy gradients for LLM training, and propose a new off-policy training objective. \citet{vassoyan2025ignore} propose ignoring KL penalties on critical tokens to boost exploration. Finally, \citet{tang2025few} study some configurations of KL estimators used in RL training of LLMs and theoretically identify biases in the estimators. However, there is no systematic study covering all the estimators used in practice, analyzing the analytical forms of the estimators and gradients as well as the downstream impact on the model performance in the on and off-policy cases. Our study fills this gap.

\section{Experimental Setup - Further Details}
\label{app:setup}
For math experiments we use the training set of Hendrycks MATH ($7500$ questions)~\citep{hendrycks2021measuring} as the training set and its test set ($5000$ questions) as the test set for reporting training behaviors. For Countdown~\citep{stojanovski2025reasoning} we generate train and held-out test sets consisting of 5000 examples each. We normalize the per-token RL objective by the number of tokens in the batch~\citep{yu2025dapo}. We run the on-policy RL fine-tuning experiments on 2 GPU nodes consisting of 4 A100s (80GB) each. For the asynchronous RL fine-tuning experiments, we fine-tune Qwen3-4B-Instruct-2507 on 1 node with 4 80GB H100 GPUs, and Qwen2.5-7B on 2 nodes having 4 80GB H100 GPUs each. We use \texttt{verl}~\citep{sheng2025hybridflow} for the on-policy RL experiments and Prime-RL~\citep{primeintellect2025prime-rl} for the asynchronous RL experiments. For evaluation, we use \texttt{lm-eval-harness}~\citep{biderman2024lessons}, using \texttt{vLLM}~\citep{kwon2023efficient} for inference with \texttt{top\_p} = 1.0, temperature = 1.0 and min\_p = 1.0.

\section{Reverse KL and Gradients for Parametric Autoregressive Model}
\label{app:toy_grad}
The closed-form expressions for the reverse KL divergence and its gradient, corresponding to parametric autoregressive model in \Cref{sec:toy} can be written as 
\begin{align}
\KL(A\|B)
&= \E_{Y\sim A}\!\Big[ \log A_\theta(Y) - \log B_\phi(Y) \Big] \\
\frac{\partial}{\partial a}\KL(A\|B)
&=  \E_{Y  \sim A}\!\Big[ \sum_{t=1}^T (y_t - p_t^A)\; (\log A_\theta(Y) - \log B_\phi(Y)) \Big],\\
\frac{\partial}{\partial b}\KL(A\|B)
&= \E_{Y \sim A}\!\Big[ \sum_{t=1}^T (y_t - p_t^A)\,c_{t-1}\; (\log A_\theta(Y) - \log B_\phi(Y)) \Big].
\end{align}

\section{Further Empirical Analysis}
In this section, we provide additional  results to further support the claims discussed in the main paper.

\subsection{Additional Evaluation Results}
\label{app:add_eval}

\begin{figure}[h]
    \centering
    \includegraphics[width=1\linewidth]{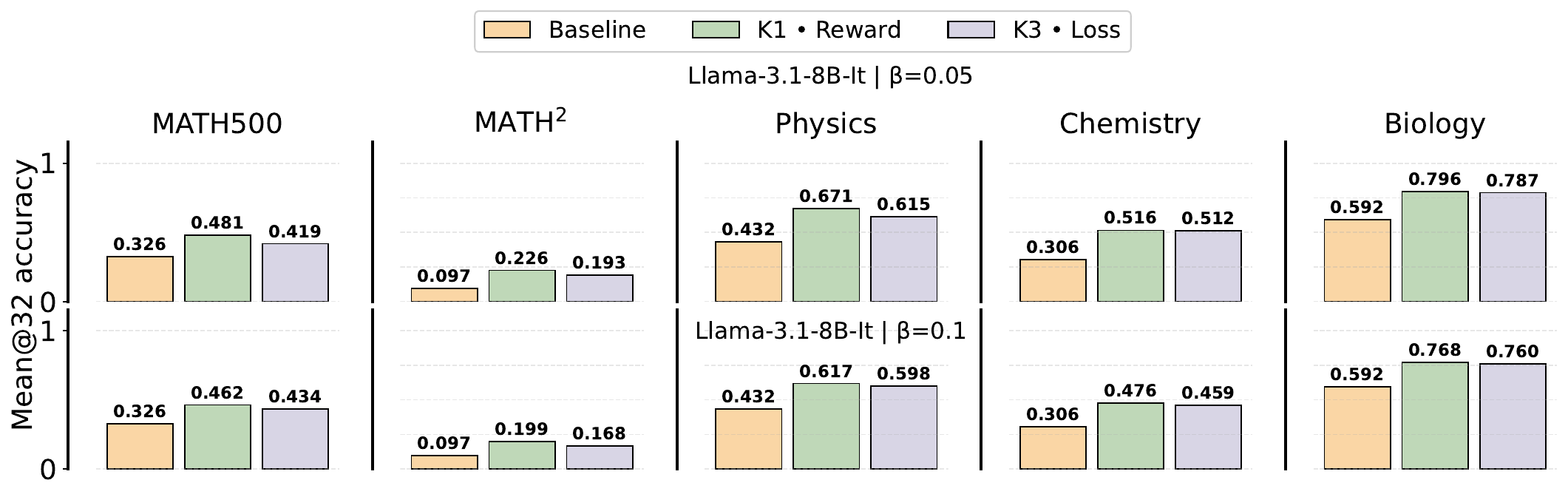}
    \vspace*{-2em}
    \caption{\textbf{Comparison of \texttt{Llama-3.1-8B-Instruct} trained with two stable estimator configurations - $\naive$-in-reward and \lowvar-in-loss.} Baseline refers to the performance of base \texttt{Llama-3.1-8B-Instruct} (prompted with chat template). $\naive$-in-reward (unbiased gradient  performs the bests on both in-domain and out-of-domain tasks. Increasing $\beta$
    deteriorates performance across the board.}
    \label{fig:ood_llama}
\end{figure}

\begin{figure}[h]
    \centering
    \includegraphics[width=\linewidth]{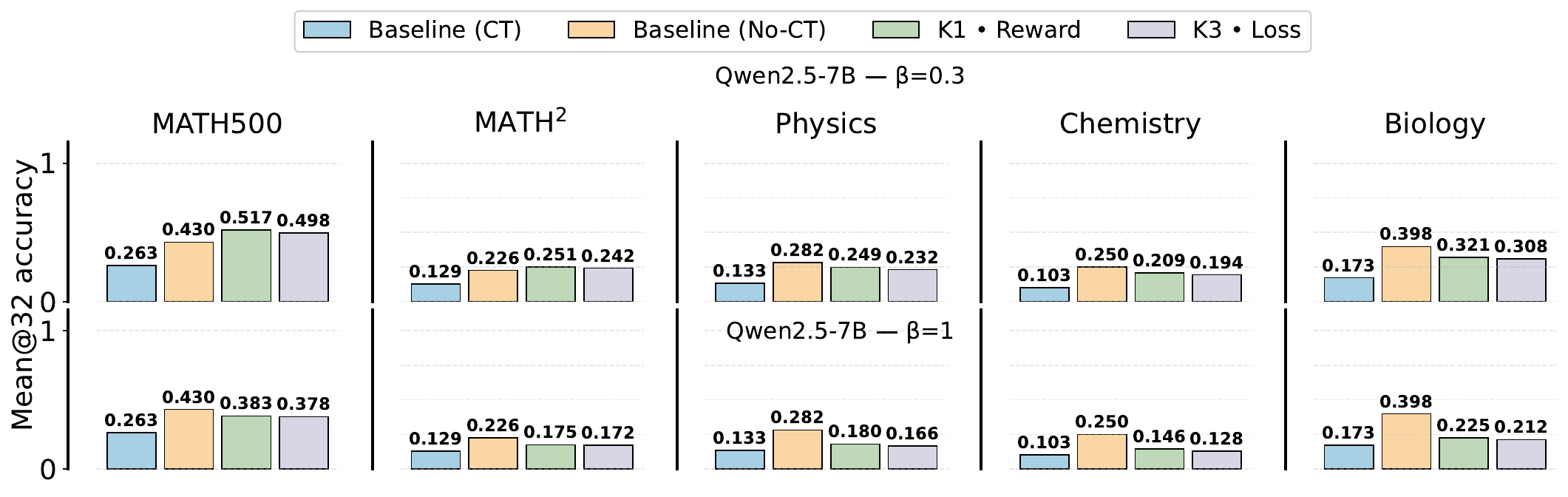}
    \caption{\textbf{Comparison of Qwen2.5-7B trained with two stable estimator configurations - $\naive$-in-reward and $\lowvar$-in-loss.} Baseline (CT) refers to the performance of base Qwen2.5-7B when prompted with a chat template. Baseline (No-CT) represents the performance when it is not prompted with a chat template. $\naive$-in-reward (unbiased gradient)  performs the best on both in-domain and out-of-domain tasks. Increasing $\beta$ consistently
    deteriorates performance.}
    \label{fig:ood_qwen_0.3_1}
\end{figure}


\subsection{Effect of Using Correct Gradient Estimators}
\label{sec:correct_results}

\begin{figure*}[h]
    \centering
    \includegraphics[width=0.25\linewidth]{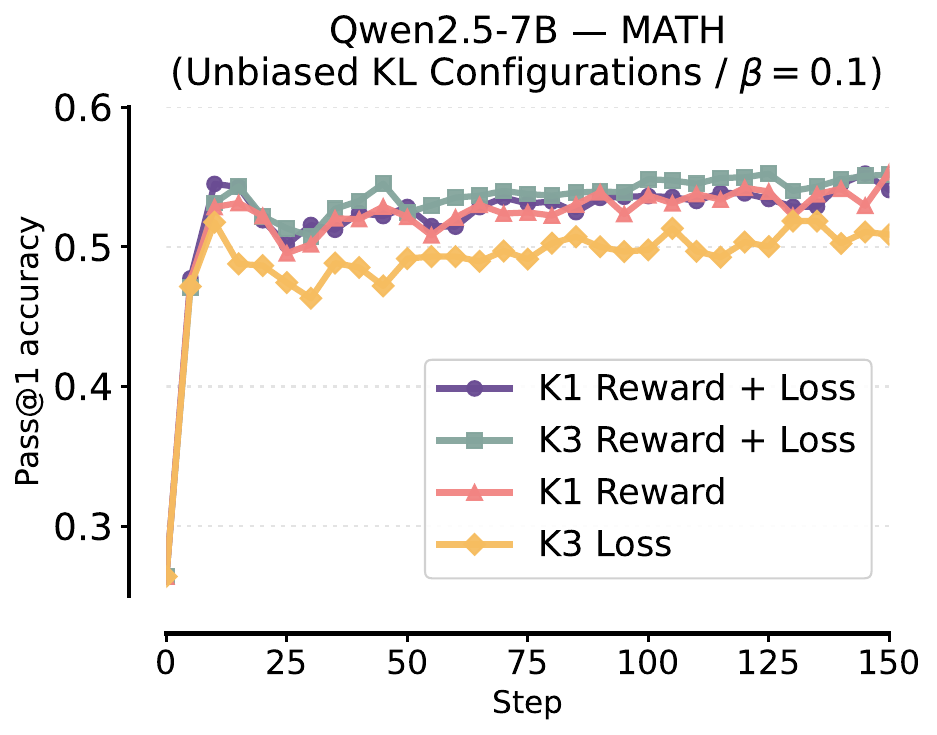}
    \includegraphics[width=0.743\linewidth]{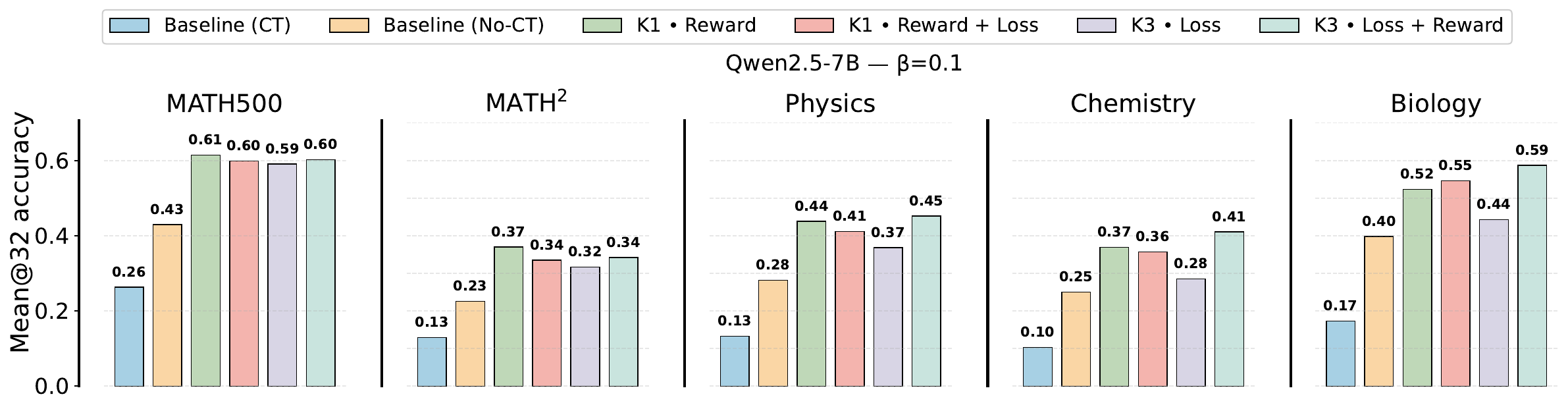}
    \caption{\textbf{Left: Train time performance of Qwen2.5-7B with different KL configurations ($\beta$ = 0.1) on MATH test.} The configurations that give unbiased gradients perform similarly, and better than $\lowvar$-in-loss. \textbf{Right: Performance of different KL configurations across several evaluation tasks.} Configurations resulting in unbiased gradients, ($\naive$-in-reward, $\naive$-in-reward-and-loss and \lowvar-in-reward-and-loss), always outperform the biased configuration (\lowvar-in-loss).}
    \label{fig:correction_eval}
\end{figure*}

As noted in \cref{sec:pos_grad}, adding the KL penalty in both reward as well as loss in on-policy settings would result in unbiased gradient estimates, regardless of the estimator. We test this empirically by training \texttt{Qwen2.5-7B} on MATH, completely on-policy ($\omega_t = 1$), using $\naive$ and \lowvar added to both reward and loss, and $\beta=0.1$ for 150 training steps with a learning rate of $10^{-6}$. All other experimental details remain the same as in \Cref{sec:results_onpolicy}. We compare the performance of these models against other estimator configurations, which are trained for the same number of steps. Results are presented in \cref{fig:correction_eval}. We observe that \naive-in-reward performs the best on in-domain tasks, whereas \lowvar added to both reward and loss performs the best on out-of-domain tasks. As a general observation, configurations inducing unbiased gradient estimates always outperform configurations inducing biased gradients, \ie, \lowvar-in-loss.

\subsection{Experiments on Code Reasoning}
\label{app:codeio}
\begin{figure}[h]
    \centering
    \includegraphics[width=0.6\linewidth]{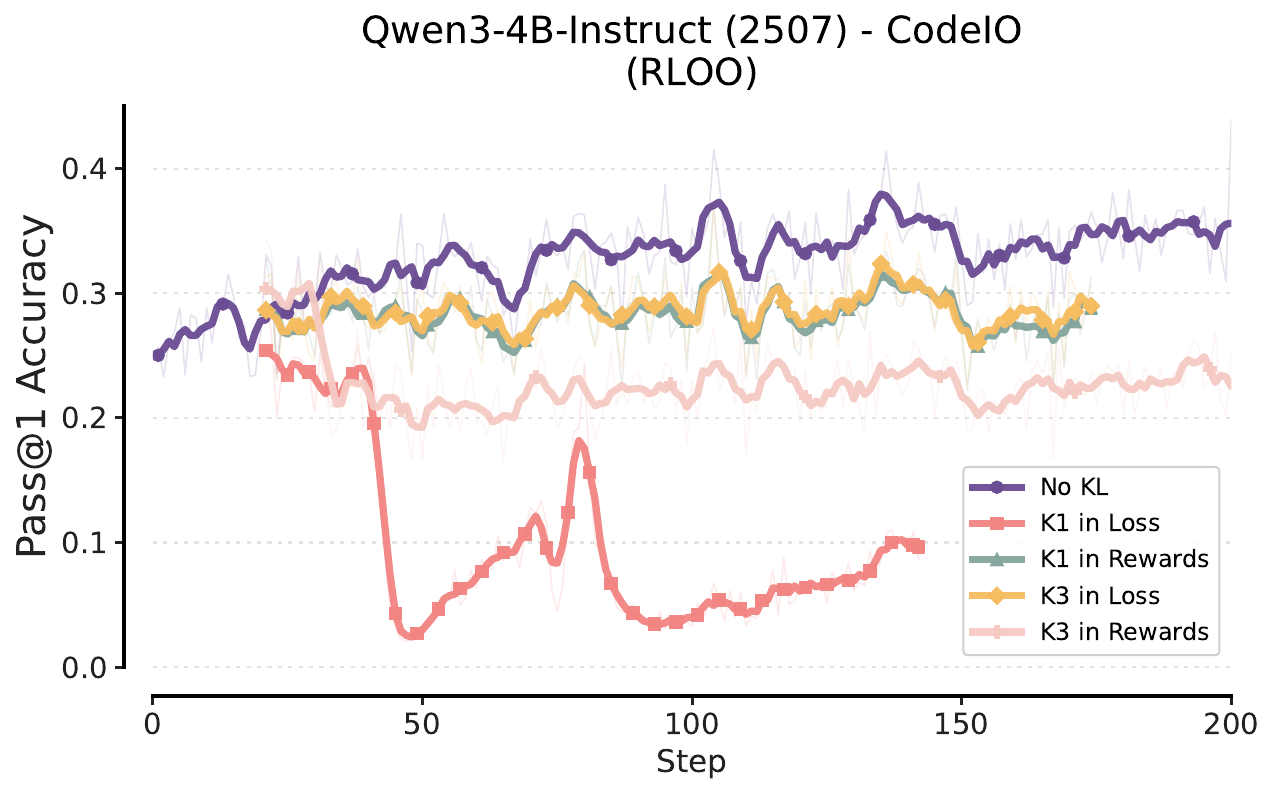}
    \vspace*{-1em}
    \caption{RL fine-tuning Qwen3-4B-Instruct-2507 on CodeIO with different KL configurations.}
    \label{fig:codeio}

\end{figure}

We validate our observation in the case of on-policy training on a different setup. We train \texttt{Qwen3-4B-Instruct-2507} on CodeIO~\citep{stojanovski2025reasoning} Given a code snippet, the model is asked to predict the outputs given a set of inputs or vice-versa, without access to a sandbox. We use a batch size of 256, maximum response budget of 4096 tokens, 4 rollouts per prompt (group size), sampling temperature of 1.0, and a fixed KL coefficient of 0.1. The learning algorithm is RLOO. We train the model on a dataset of 15000 procedurally generated problems and evaluate the model on a held out test set. The test performance over the course of training is shown in ~\Cref{fig:codeio}. The observations in code reasoning remain consistent with the results in the main text: No KL performs best in this policy setting, while K1 in loss and K3 in reward tend to hurt performance.

\subsection{Further Results with Off-Policy Training}
\begin{figure*}[t]
    \centering
    \includegraphics[width=0.6\linewidth]{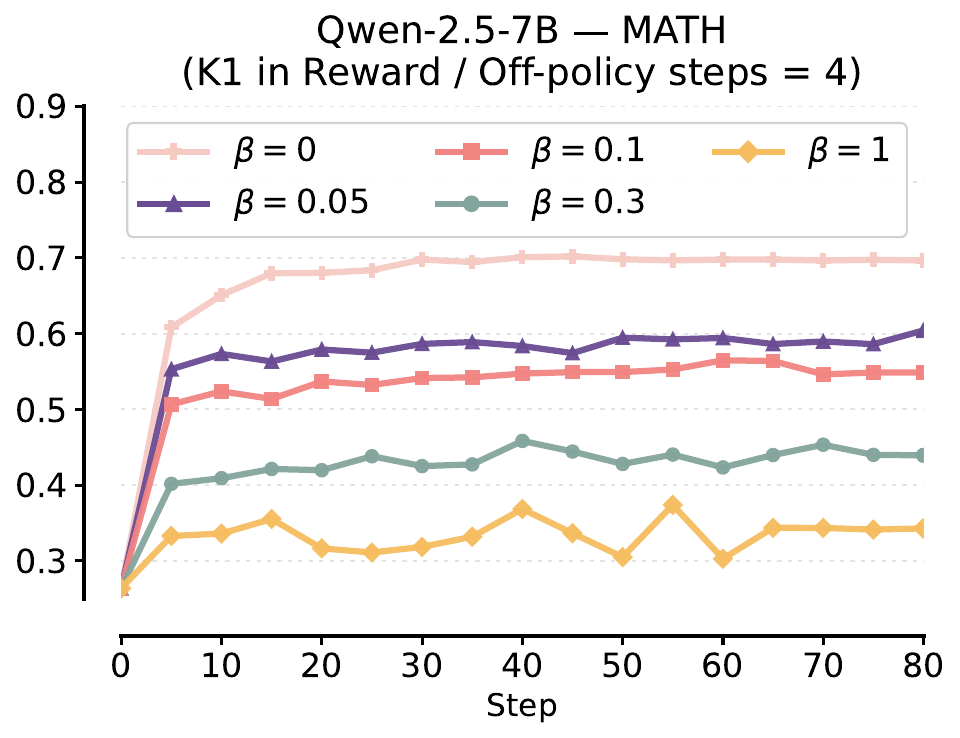}
    \vspace*{-1em}
    \caption{\textbf{RL fine-tuning Qwen2.5-7B with \naive-in-reward, total training batch size of 1024 and mini-batch size of 256.} The training remains stable as opposed to configurations leading to biased gradients which result where the instabilities are accentuated as compared to on-policy training.}
    \label{fig:k1r_offpolicy}

\end{figure*}

Figure~\ref{fig:k1r_offpolicy} shows the performance of Qwen2.5-7B on MATH test set, while being fine-tuned on MATH training data, with a training batch size of 1024 and a mini-batch size of 256, i.e. $\omega \neq 1$, using \naive-in-reward. Despite the off-policy, the training is stable as opposed to \naive-in-loss and \lowvar-in-reward with $\omega \neq 1$, where the training instabilities are accentuated as compared to the on-policy case.

\subsection{Further Analysis for \naive-in-reward and \lowvar-in-loss}
\label{app:analysis}

\begin{figure*}[t]
    \centering
    \includegraphics[width=0.49\linewidth]{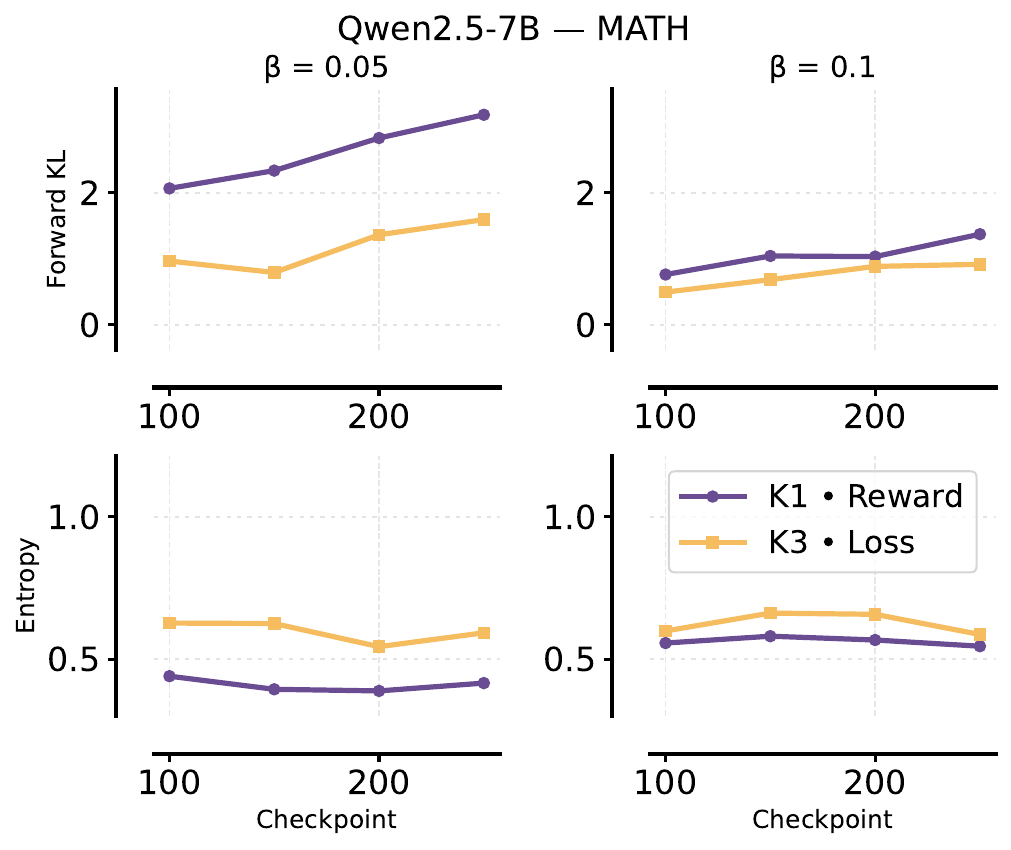}
    \includegraphics[width=0.49\linewidth]{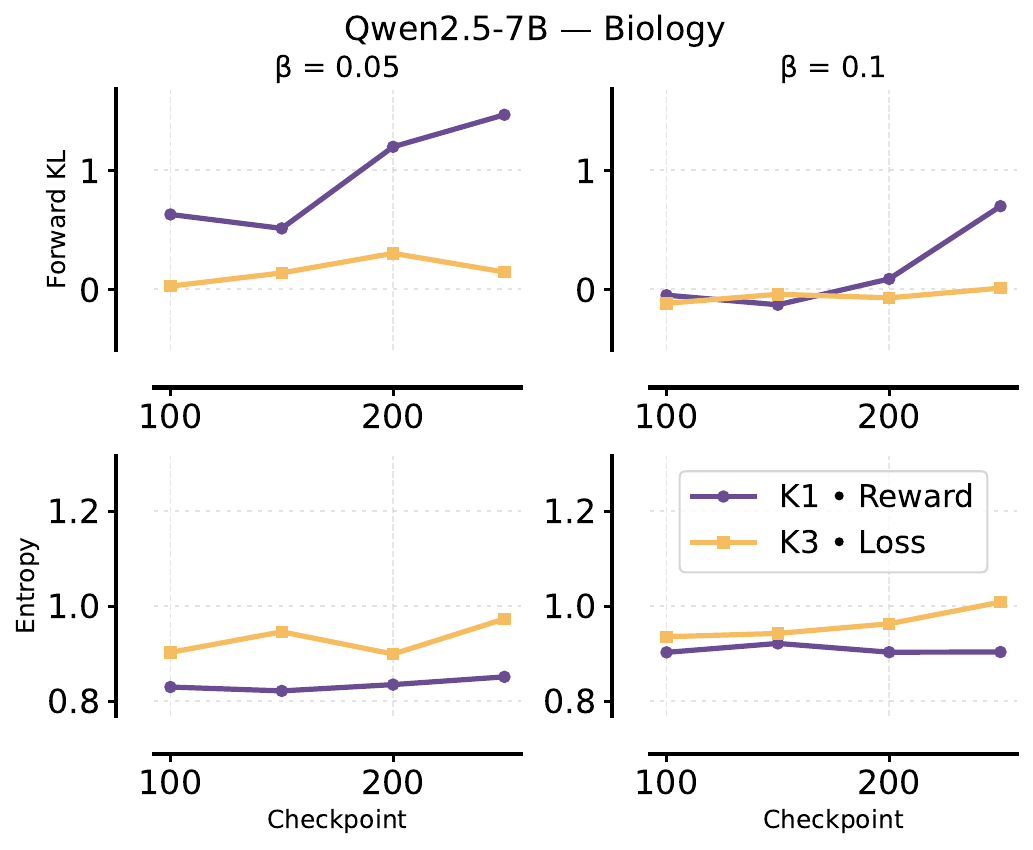}
    \vspace*{-1em}
    \caption{Sequence level forward KL with respect to the base policy and entropy across different training checkpoints, computed on MATH test dataset and MMLU Biology dataset}
    \label{fig:analysis}
\end{figure*}

In an attempt to understand the performance difference between \lowvar-in-loss and \naive-in-reward, we plot two metrics: (1) the sequence-level forward KL divergence between the reference policy $\piref$ and the training policy $\pith$, and (2) the entropy of the training policy $\pith$. We compute these metrics for different checkpoints of Qwen2.5-7B while being trained on MATH, on one in-distribution task - MATH test dataset and one out-of-domain task - MMLU Biology. The results are shown in Figure~\ref{fig:analysis}. We observe that the forward KL divergence remains lower in the case of \lowvar-in-loss whereas entropy stays lower for \naive-in-reward. However, these observations do not directly explain the performance differences between \naive-in-reward and \lowvar-in-loss and further analysis needs to be carried out.

\section{K1 in Reward in Different Libraries}
Table~\ref{tab:kl_config} lists the arguments to be included while submitting the training scripts to use the unbiased K1 in reward configuration discussed in this work for three popular libraries - VeRL~\citep{sheng2025hybridflow}, OpenRLHF~\citep{hu2024openrlhf} and SkyRL~\citep{cao2025skyrl}. Another popular library under consideration - Prime-RL~\citep{primeintellect2025prime-rl} seems to have removed the previously used KL in loss term entirely.

\begin{table}[h!]
\label{tab:lib_settings}
\centering
\tiny
\begin{tabular}{ll}
\toprule
\textbf{Library} & \textbf{Arguments} \\
\midrule
VeRL 
& \texttt{algorithm.use\_kl\_in\_reward=True, algorithm.kl\_penalty="kl"} \\
OpenRLHF 
& \textbf{DO NOT} set \texttt{--use\_kl\_loss, --kl\_estimator="k1"} \\
SkyRL
& \texttt{trainer.algorithm.kl\_estimator\_type="k1", trainer.algorithm.use\_kl\_in\_reward=True} \\
\bottomrule
\end{tabular}
\caption{KL regularization configuration across different libraries.}
\label{tab:kl_config}
\end{table}

\section{Mathematical Details and Derivations}
\label{app:math}

\textbf{Notation.} We first define the notation used for the analysis into the bias of the estimators and their corresponding gradients.

\begin{table}[h!]
\centering
\begin{tabular}{ll}
\toprule
\textbf{Symbol} & \textbf{Description} \\
\midrule
$\pith$  & policy trained using RL\\
$\piref$ & reference policy \\
$x$ & prompt \\
$\yseq$ & generated response with $T$ tokens \\
$y_t$ & token at position $t$ of response $\yseq$ \\
\naive & naïve estimator of $D_\text{KL}(\pith || \piref)$\\
\lowvar & Schulman estimator of $D_\text{KL}(\pith || \piref)$\\
$\naivet$ & naïve estimator of $D_\text{KL}(\pith || \piref)$ at token $t$\\
$\lowvart$ & Schulman estimator of $D_\text{KL}(\pith || \piref)$ at token $t$\\
\bottomrule
\end{tabular}
\caption{Notation table.}
\label{tab:notation}
\end{table}

\subsection{True Gradient}
\label{app:true_grad}
We want to estimate the gradient of the KL divergence between $\pith$ and $\piref$ that are defined over entire sequences of tokens. Specifically, we want:
\begin{align}
    \nabla_\theta \kl(\pith(\cdot \mid x) \,\|\, \piref(\cdot \mid x)) &= \nabla_\theta \mathbb{E}_{\yseq \sim \pith(\cdot \mid x)}\left[\log \frac{\pith(\yseq \mid x)}{\piref(\yseq \mid x)}\right]\\
    &= \mathbb{E}_{\yseq \sim \pith(\cdot \mid x)}\left[\log \frac{\pith(\yseq \mid x)}{\piref(\yseq \mid x)}\,\nabla_\theta \log \pith(\yseq \mid x)\right].
    \label{eq:kl-seq-grad}
\end{align}
This is the true \emph{sequence}-level gradient of the KL divergence. Every gradient estimator we use henceforth aims to estimate this true gradient.

\subsection{Path-wise and Score Function Derivatives}
\label{app:est_grad}
We show how the gradient of the KL estimator, or any other function, decomposes into a \emph{path-wise} derivative corresponding to the gradient of the estimator inside the expectation, and the \emph{score function} derivative arising from the $\theta$-dependent sampling in the expectation.
\begin{align}
    &\d \expec\left[\widehat{\mathrm{KL}}\right] \\
    &\qquad= \d \sum_{\yseq} \widehat{\mathrm{KL}} \cdot \pith(\yseq \mid x) \\
    &\qquad= \sum_{\yseq} \left(\d \widehat{\mathrm{KL}}\right) \cdot \pith(\yseq \mid x) + \sum_{\yseq}\widehat{\mathrm{KL}} \cdot \left(\d \pith(\yseq \mid x)\right) \\
    &\qquad= \underbrace{\mathbb{E}_{\yseq \sim \pith(\cdot \mid x)} \d \left[\widehat{\mathrm{KL}}\right]}_{\text{path-wise derivative}} + \underbrace{\mathbb{E}_{\yseq \sim \pith(\cdot \mid x)} \left[\widehat{\mathrm{KL}} \cdot \d \log \pith(\yseq \mid x)\right]}_{\text{score function derivative}} \\
    &\qquad= \underbrace{\mathbb{E}_{\yseq \sim \pith(\cdot \mid x)} \left[\sum_t \d\widehat{\mathrm{KL}}_t\right]}_{\text{Pathwise}} + \underbrace{\mathbb{E}_{\yseq \sim \pith(\cdot \mid x)} \left[\left(\sum_t\widehat{\mathrm{KL}}_t\right) \cdot \d \log \pith(\yseq \mid x)\right]}_{\text{Score function}},
\end{align}
where in the last line we write the KL divergence estimator as the sum of estimators at each individual token. Note that the path-wise derivative corresponds to using the estimator directly in the loss (and backpropagating through it), whereas the score function derivative corresponds to adding the estimator to the reward.

\subsection{$\naive$ Estimator}
\label{app:k1}
The $\naive$ estimator for a sequence $\yseq$ can be written as:
\begin{align}
    \naive &= \sum_{t=1}^T \naivet = \sum_{t=1}^T \log \frac{\pith(y_t \mid x,y_{<t})}{\piref(y_t \mid x,y_{<t})}.
\end{align}
It is easy to see that this is an unbiased estimator of $D_\text{KL}(\pith || \piref)$.

To analyze the gradient of \naive, we calculate the path-wise derivative and the score function derivative separately. 

\paragraph{Path-wise derivative.} 
The path-wise derivative of $\naive$ evaluates to zero under expectation.
\begin{align}
    &\mathbb{E}_{\yseq \sim \pith(\cdot \mid x)}\left[\d \sum_t \naivet\right] = 0.
\end{align}
\paragraph{Score function derivative.} The score function derivative of \naive is an unbiased estimate of the true gradient in \Cref{eq:kl-seq-grad}.
\begin{align}
    &\mathbb{E}_{\yseq \sim \pith(\cdot \mid x)}\left[\left(\sum_t \naivet\right)\cdot \d \log \pith(\yseq \mid x)\right] \nonumber \\
    &\qquad = \mathbb{E}_{\yseq \sim \pith(\cdot \mid x)}\left[\log \frac{\pith(\yseq \mid x)}{\piref(\yseq \mid x)}\,\d \log \pith(\yseq \mid x)\right].
\end{align}
Therefore, adding $\naive$ estimator to the reward results in an unbiased estimate of the gradient of the KL-regularized RL objective, whereas using $\naive$ in the loss directly does not. In fact, since the path-wise derivative of $\naive$ is zero in expectation, in principle, using it in the loss should be equivalent to optimizing the RL objective without KL regularization (\ie, $\beta = 0$). In practice, however, using this term in the loss introduces some variance that can hurt the optimization. We also note that we can reduce the variance of the score function derivative by removing the past tokens from the inner sum, since their contribution to the gradient will be zero in expectation.

\begin{tcolorbox}[colback=orange!10,
leftrule=0.5mm,top=1mm,bottom=1mm,boxrule=0.6pt]
\textbf{Takeaway for \naive:} 
\begin{itemize}[left=0cm]
    \item Adding $\naive$ to the reward gives us an unbiased estimate of the gradient of the KL-regularized RL objective.
    \item Using $\naive$ in loss results in a biased estimate of the true gradient and is equivalent to using no KL regularization, but can introduce some variance in practice.
\end{itemize}
\end{tcolorbox}


\subsection{$\lowvar$ Estimator}
\label{app:k3}

The \lowvar estimator for a sequence $\yseq$ can be written as:
\begin{align}
    \lowvar = \sum_{t=1}^T \lowvart = \sum_{t=1}^T \left(\frac{\piref(y_t \mid y_{<t}, x)}{\pith(y_t \mid y_{<t}, x)} - 1 - \log \frac{\piref(y_t \mid y_{<t}, x)}{\pith(y_t \mid y_{<t}, x)}\right).
\end{align}
We first show that \lowvar is an unbiased estimator of $D_\mathrm{KL}(\pith \,\|\, \piref)$:
\begin{align}
    &\mathbb{E}_{\yseq \sim \pith(\cdot\mid x)}\left[\lowvar\right] \\
    &\qquad = \mathbb{E}_{\yseq \sim \pith(\cdot\mid x)}\left[\sum_t \left(\frac{\piref(y_t \mid x, y_{<t})}{\pith(y_t \mid x, y_{<t})} - 1 - \log \frac{\piref(y_t \mid x, y_{<t})}{\pith(y_t \mid x, y_{<t})}\right)\right] \\
    &\qquad = \mathbb{E}_{\yseq \sim \pith(\cdot\mid x)}\left[\sum_t \frac{\piref(y_t \mid x, y_{<t})}{\pith(y_t \mid x, y_{<t})}\right] - T + \mathbb{E}_{\yseq \sim \pith}\left[\log \frac{\pith(\yseq \mid x)}{\piref(\yseq \mid x)}\right] \\
    &\qquad = D_\mathrm{KL}(\pith \,\|\, \piref)
\end{align}
We again calculate the path-wise and the score function derivatives of \lowvar separately.
\paragraph{Path-wise derivative.} The path-wise derivative of \lowvar is a biased estimate of the true gradient in \Cref{eq:kl-seq-grad}.
\begin{align}
    &\mathbb{E}_{\yseq \sim \pith(\cdot \mid x)}\left[\d \lowvart\right] \\
    &\qquad= \mathbb{E}_{\yseq \sim \pith(\cdot \mid x)}\left[\d \sum_t \left(\frac{\piref(y_t \mid x, y_{<t})}{\pith(y_t \mid x, y_{<t})} - 1 - \log \frac{\piref(y_t \mid x, y_{<t})}{\pith(y_t \mid x, y_{<t})}\right)\right] \\
    &\qquad= -\mathbb{E}_{\yseq \sim \pith(\cdot \mid x)}\left[\sum_t\frac{\piref(y_t \mid x, y_{<t})}{\pith(y_t \mid x, y_{<t})} \d \log \pith(y_t \mid x, y_{<t}) \right] \nonumber + \mathbb{E}_{\yseq \sim \pith(\cdot \mid x)}\left[\d \texttt{K1}\right] \\
    &\qquad= - \sum_t\mathbb{E}_{y_{<t} \sim \pith(\cdot \mid x)}\mathbb{E}_{y_t \sim \piref(\cdot \mid x, y_{<t})}\left[\d \log \pith(y_t \mid x, y_{<t})\right] \\
    &\qquad= \sum_t \mathbb{E}_{y_{<t} \sim \pith(\cdot \mid x)} \left[\d \mathrm{KL}(\piref(\cdot \mid x, y_{<t}) \,\|\, \pith(\cdot \mid x, y_{<t}))\right].
\end{align}
The above expression resembles the gradient of the \emph{forward} KL divergence at the token level, except that the samples are drawn from $\pith(\cdot | x)$ instead of $\piref(\cdot | x)$.
\paragraph{Score function derivative.} The score function derivative of \lowvar also is a biased estimate of the true gradient in \Cref{eq:kl-seq-grad}.
\begin{align}
    &\mathbb{E}_{\yseq \sim \pith(\cdot \mid x)}\left[\left(\sum_{t=1}^T\lowvart\right)\cdot \d \log \pith(\yseq \mid x)\right] \nonumber \\
    &\qquad = \mathbb{E}_{\yseq \sim \pith(\cdot \mid x)}\left[\left(\sum_t \left(\frac{\piref(y_t \mid x, y_{<t})}{\pith(y_t \mid x, y_{<t})} - 1 - \log \frac{\piref(y_t \mid x, y_{<t})}{\pith(y_t \mid x, y_{<t})}\right)\right)\d \log \pith(\yseq \mid x)\right] \\
    &\qquad= \mathbb{E}_{y_{1:T} \sim \pith(\cdot \mid x)}\left[\sum_t \frac{\piref(y_t \mid x, y_{<t})}{\pith(y_t \mid x, y_{<t})}\d \log \pith(y_{1:T} \mid x)\right] \nonumber \\&\qquad\qquad+ \mathbb{E}_{\yseq \sim \pith(\cdot \mid x)}\left[\log \frac{\pith(\yseq \mid x)}{\piref(\yseq \mid x)}\d \log \pith(\yseq \mid x)\right] \\
    &\qquad= \mathbb{E}_{y_{1:T} \sim \pith(\cdot \mid x)}\left[\sum_t \sum_s \frac{\piref(y_t \mid x, y_{<t})}{\pith(y_t \mid x, y_{<t})}\d \log \pith(y_{s} \mid y_{<s}, x)\right] \nonumber \\&\qquad\qquad+ \mathbb{E}_{\yseq \sim \pith(\cdot \mid x)}\left[\log \frac{\pith(\yseq \mid x)}{\piref(\yseq \mid x)}\d \log \pith(\yseq \mid x)\right] \label{eq:low_var_kl_cancel} \\
    &= - \sum_t\mathbb{E}_{y_{<t} \sim \pith(\cdot \mid x)}\left[\d \mathrm{KL}(\piref(\cdot \mid x, y_{<t}) \,\|\, \pith(\cdot \mid x, y_{<t}))\right] + \mathbb{E}_{\yseq \sim \pith(\cdot \mid x)}\left[\texttt{K1}\cdot \d \log \pith(\yseq \mid x)\right]. \label{eq:low_var_kl_final}
\end{align}
where in Equation \ref{eq:low_var_kl_cancel} the terms corresponding to $s < t$ and $s > t$ reduce to $0$.
The first term in \Cref{eq:low_var_kl_final} represents the bias with respect to the true gradient. Therefore, using \lowvar either in loss or added to the reward results in a biased estimate of the true gradient. 

We note that the path-wise derivative of \lowvar corresponds to the regularization term used in GRPO \citep{shao2024deepseekmath}, a popular RL algorithm used for training LLMs.\\

\begin{tcolorbox}[colback=orange!10,
leftrule=0.5mm,top=1mm,bottom=1mm,boxrule=0.6pt]
\textbf{Takeaway for \lowvar:} 
Adding \lowvar to the reward or using it in the loss results in a biased estimate of the gradient of the KL-regularized RL objective.
\end{tcolorbox}

\end{document}